%% file: main.tex
\documentclass[lettersize,journal]{IEEEtran}
\usepackage{amsmath,amsfonts}
\usepackage{algorithmic}
\usepackage{array}
\usepackage{textcomp}
\usepackage{stfloats}
\usepackage{url}
\usepackage{verbatim}
\usepackage{graphicx}
\usepackage{cite}
\usepackage{enumerate}

\usepackage{amsmath}
\DeclareMathOperator*{\argmax}{argmax}

\newcommand{\stitle}[1]{\vspace{1ex} \noindent{\bf #1}}
\usepackage{bbding} 
\usepackage{pifont}
\usepackage{tablefootnote}
\usepackage{hyperref}
\newtheorem{definition}{Definition}
\usepackage{bigstrut}

\usepackage{titlesec}
\usepackage{graphicx}
\usepackage{svg}
\graphicspath{ {image/} }  
\usepackage{amssymb}
\usepackage{booktabs} 
\usepackage{bbding} 
\usepackage{amsmath}
\usepackage{amsfonts}
\usepackage{array}   
\usepackage{tabularx} 
\usepackage{booktabs} 
\usepackage[linesnumbered,ruled,vlined]{algorithm2e}
\usepackage{svg}
\usepackage{multicol}
\usepackage{multirow}
\usepackage{adjustbox}
\usepackage[most]{tcolorbox}
\usepackage{subcaption}
\usepackage{xcolor}
\newtcolorbox{answerbox}{
  colback=gray!10, 
  colframe=gray!50, 
  sharp corners,
  boxrule=0.5pt,
  left=4pt,
  right=4pt,
  top=4pt,
  bottom=4pt,
  breakable
}
\usepackage{balance}

\definecolor{mygreen}{HTML}{02818a}

\def\BibTeX{{\rm B\kern-.05em{\sc i\kern-.025em b}\kern-.08em
    T\kern-.1667em\lower.7ex\hbox{E}\kern-.125emX}}
\usepackage{balance}
\begin{document}
\title{Robust Smart Contract Vulnerability Detection via Contrastive Learning-Enhanced Granular-ball Training}

\author{Zeli Wang, Qingxuan Yang, Shuyin Xia, Yueming Wu, Bo Liu, Longlong Lin 
\IEEEcompsocitemizethanks{\IEEEcompsocthanksitem Zeli Wang, Qingxuan Yang, Shuyin Xia are with Chongqing Key Laboratory of Computational Intelligence, Key Laboratory of Cyberspace Big Data Intelligent Security, Ministry of Education, 
and Key Laboratory of Big Data Intelligent Computing, Chongqing University of Posts and Telecommunications. 
Email: zlwang@cqupt.edu.cn, s230231148@stu.cqupt.edu.cn, xiasy@cqupt.edu.cn;
\IEEEcompsocthanksitem Yueming Wu is with the School of Cyber Science and Engineering, Huazhong University of Science and Technology. Email: yuemingwu@hust.edu.cn;
\IEEEcompsocthanksitem Bo Liu is with School of Computer Science and Artificial Intelligence, Zhengzhou University, China. Email: liubo@zzu.edu.cn;
\IEEEcompsocthanksitem Longlong Lin is with the College of Computer and Information Science, Southwest University, China. Email: longlonglin@swu.edu.cn.

        }
	\thanks{Manuscript received XXX, XXX; revised XXX, XXXX.}}

\markboth{Journal of \LaTeX\ Class Files,~Vol.~18, No.~9, September~2020}%
{How to Use the IEEEtran \LaTeX \ Templates}

\maketitle
\begin{abstract}
  Blockchain has become the fundamental infrastructure for building a digital economy society. As the key technology to implement Blockchain applications, smart contracts face serious security vulnerabilities. Deep neural networks (DNNs) have emerged as a prominent approach for detecting smart contract vulnerabilities, driven by the growing contract datasets and advanced deep learning techniques. However, DNNs typically require large-scale labeled datasets to model the relationships between contract features and vulnerability labels. In practice, the labeling process often depends on existing open-sourced tools, whose accuracy cannot be guaranteed. Consequently, label noise poses a significant challenge for the accuracy and robustness of the smart contract, which is rarely explored in the literature. To this end, we propose \underline{C}ontrastive learning-enhanced \underline{G}ranular-\underline{B}all smart \underline{C}ontracts training, CGBC, to enhance the robustness of contract vulnerability detection. Specifically, CGBC first introduces a Granular-ball computing layer between the encoder layer and the classifier layer, to group similar contracts into Granular-Balls (GBs) and generate new coarse-grained representations (i.e., the center and the label of GBs) for them, which can correct noisy labels based on the most correct samples. An inter-GB compactness loss and an intra-GB looseness loss are combined to enhance the effectiveness of clustering. Then, to improve the accuracy of GBs, we pretrain the model through unsupervised contrastive learning supported by our novel semantic-consistent smart contract augmentation method. This procedure can discriminate contracts with different labels by dragging the representation of similar contracts closer, assisting CGBC in clustering. Subsequently, we leverage the symmetric cross-entropy loss function to measure the model quality, which can combat the label noise in gradient computations. Finally, extensive experiments show that the proposed CGBC can significantly improve the robustness and effectiveness of the smart contract vulnerability detection under various noisy settings when contrasted with baselines.
\end{abstract}


\section{Introduction}
\IEEEPARstart{B}{lockchain} is a distributed and decentralized ledger technology, which has been widely applied across various domains \cite{wang2021ethna, liu2023blockchain}. Smart contracts, which serve as the core mechanism for automating and enforcing business logic on blockchain platforms, have become indispensable for enabling these applications. However, the prevalence of security vulnerabilities in smart contracts remains a major obstacle, significantly hindering the broader adoption and real-world deployment of blockchain-based solutions~\cite{10700860, DBLP:journals/tse/JinWWDZZ22, 8967006}.
Vulnerability detection is a widely studied approach to enhancing smart contract security. Traditional program analysis methods often depend on manually defined vulnerability patterns, which are constrained by expert knowledge and struggle to scale with the intricate logic semantics of smart contracts \cite{DBLP:journals/jnca/WangDLCZ24,EVulHunter,DetectAb}. 
Deep learning offers a promising solution by automatically learning vulnerability patterns through its powerful pattern recognition capabilities \cite{dlva,Cross-Modality,rlrep,clear,MTVHunter}. 

MTVHunter leverages multiple teacher models to guide the training process by aggregating their predictions into more accurate soft labels, improving robustness and accuracy \cite{MTVHunter}. Clear achieves significant detection accuracy improvements by generating labeled contract pairs through data sampling and extracting semantic features using a Transformer module, followed by model fine-tuning for vulnerability detection \cite{clear}. 
Although these learning-based models have achieved remarkable performance in detecting smart contract vulnerabilities, their effectiveness relies on extensive high-quality labeled datasets. In practice, existing smart contract datasets are generally annotated by open-source tools \cite{dlva,Cross-Modality,clear}
However, the accuracy of current contract vulnerability detection tools is far from expectations: for example, the empirical study on 47,587 contracts and nine tools demonstrate that only 42\% of contract vulnerabilities are detected \cite{DBLP:conf/icse/DurieuxFAC20}; the research conducted on 18 contract detection tools proves that they have poor performance. 
Consequently, the existing tool-based label procedure will inevitably introduce much label noise, which can significantly degrade model performance \cite{10646883}. Thus, mitigating the impact of label noise on learning-based vulnerability detection is urgent.

Much research has been conducted to resolve the label noise in computer vision \cite{pico,PNT} and natural language processing \cite{GBRain,Defense}, but only a few studies explore this problem in software vulnerability detection. 
ACGVD \cite{ACGVD} adopts a framework that combines GNN and GAT, constructing a comprehensive graph representation that integrates various types of information from source code. This approach mitigates the noise interference that may arise from relying on a single information source, enhancing the model's ability to adapt to complex data. \cite{ReGVD} designs adaptive graph convolution operations that allow the model to adjust to noisy patterns automatically, and employs a weighted loss function to mitigate the impact of noisy labels. These tools demonstrate that deep learning-based vulnerability detection methods in traditional software systems are susceptible to label noise and propose effective mitigation strategies. However, these approaches cannot be directly generalized to smart contracts due to their fundamental differences, such as programming languages and logical architectures. To the best of our knowledge, no countermeasures have been proposed to address label noise in smart contract vulnerability detection frameworks.

An intuitive way to address this problem is to identify and correct noisy labels. However, it is nontrivial and computationally exhaustive to spot all noisy labels precisely. To alleviate this dilemma, bearing the core observation that most samples are correctly annotated, we aim to cluster similar smart contracts and remove noise through the centers of clusters, which represent the common features and most labels. There are two key challenges to achieve this goal: (1) generating close embeddings for similar contracts. The embeddings of smart contracts determine their similarity, namely, the correctness of clustering. Since smart contracts have complicated code structures and diverse user-defined variables, it is tricky to embed contracts with the same labels closely. (2) Adaptively clustering smart contracts. Smart contracts even belonging to the same category may have distinct representative features. This implies that smart contracts should be clustered according to their data distribution features, rather than strictly clustering the contracts with the same label into a set. Such requirements ask for an adaptive multi-granularity clustering approach.

To overcome these challenges, we propose a novel robust training approach for learning-based smart contract detection models through contrastive learning-enhanced Granular-ball computing (CGBC). Our CGBC achieves coarse-grained representation learning through a Granular-ball (GB) clustering layer. The label and vector representation of a GB center are the new coarse-grained representation for all samples in the current GB. This procedure can correct noisy labels and diminish their influence. To address the first challenge and bring similar contracts closer, we pretrain CGBC with unsupervised contrastive learning to optimize the embeddings of smart contracts. Additionally, we introduce the intra-GB compactness loss and inter-GB looseness loss function to the fine-tuning loss function to further pull the embeddings of similar contracts closer. For the second challenge, we adopt the Granular-ball computing method, an adaptively multi-granularity clustering algorithm, to automatically group smart contracts following data distribution features, without predefined parameters such as the amount or size of sets. By integrating these designs, CGBC realizes robust vulnerability detection for smart contracts.
We conduct extensive experiments on a large-scale dataset across three common vulnerability types: Reentrancy (RE), Timestamp Dependency (TD), and Integer Overflow (IO). The results demonstrate that CGBC significantly improves the robustness of smart contract vulnerability detection under label noise. Even with 40\% noisy labels, CGBC maintains high performance with an average F1-score of 76.59\%, outperforming all baseline methods. These findings confirm the effectiveness and robustness of CGBC in addressing label noise, enhancing the reliability of smart contract analysis. 
To sum up, the contributions of this paper are listed as follows:

\begin{itemize}
    \item We present the first contrastive learning-enhanced Granular-ball training method for robust learning-based smart contract vulnerability detection under noisy environments, which leverages unsupervised pretraining to initialize contract representations, and then uses coarse-grained representation learning to mitigate label noise in vulnerability detection.
    \item We propose a semantic-preserving smart contract augmentation to conduct contrastive learning. Furthermore, we introduce a Granular-ball-based coarse-fine robust training method, assisted by intra-GB and extra-GB quality metrics for Granular-balls, to conduct robust learning for smart contract vulnerability detection.
    \item Experimental results on abundant evaluations demonstrate the effectiveness and robustness of our proposal. Moreover, our proposal always outperforms state-of-the-art models. 
\end{itemize}

\begin{table*}[t]
\centering
\caption{Comparison of representative related studies and our method.}
\label{tab:related_summary}
\renewcommand{\arraystretch}{1.08}
\setlength{\tabcolsep}{3pt}
\scriptsize
\begin{tabular}{p{0.9cm} p{7cm} p{7cm}}
\toprule
\textbf{Ref.} & \textbf{Main work} & \textbf{Remarks} \\
\midrule
\cite{oyente} & Symbolic execution is used to explore contract execution paths for vulnerability detection. & It gives clear semantic analysis, but it suffers from path explosion and high analysis cost. \\
\midrule
\cite{slither} & Static analysis is performed by extracting structural and dependency information from smart contracts. & It is practical and efficient, but it still relies heavily on handcrafted rules. \\
\midrule
\cite{Verx} & Formal verification is used to check safety properties of smart contracts. & It provides stronger theoretical support, but it needs formal modeling and has high verification cost. \\
\midrule
\cite{EthPloit} & Fuzzing is combined with taint analysis and feedback guidance to generate exploit-oriented test cases. & It can better validate triggerable vulnerabilities than pure static analysis, but it is still sensitive to seed quality and path coverage. \\
\midrule
\cite{AME} & Graph neural networks are combined with interpretable graph features and expert patterns for smart contract vulnerability detection. & It balances representation learning and interpretability, but its flexibility is limited by expert-designed knowledge. \\
\midrule
\cite{10700860} & Vulnerability subgraphs are constructed and graph neural networks are used for Ethereum smart contract vulnerability detection. & It improves vulnerability-oriented structural modeling, but it still assumes relatively reliable labels during training. \\
\midrule
\cite{clear} & Contrastive learning is introduced to improve the discriminative quality of smart contract representations. & It improves representation learning, but it does not explicitly handle robust training under noisy supervision. \\
\midrule
\cite{Nie} & Confidence learning and differential training are proposed to identify suspicious labels in vulnerability datasets. & It shows the importance of noisy-label robustness, but it mainly focuses on general software vulnerability detection. \\
\midrule
\textbf{Ours} & Contrastive learning and granular-ball-based robust training are jointly introduced for smart contract vulnerability detection. & It improves both representation quality and robustness to noisy supervision in one framework. \\
\bottomrule
\end{tabular}
\end{table*}

\section{Related Work}
This section reviews prior studies on smart contract vulnerability detection. Existing work can be classified into four main categories, namely program analysis, deep learning, large language models, and robust vulnerability detection. To make the comparison clearer, Table~\ref{tab:related_summary} summarizes representative studies and our method in terms of main work and remarks.

\subsection{\textbf{Program Analysis Method}}
Program analysis are classifical methods to detect smart contract vulnerabilities. They are still widely used because they provide clear semantic reasoning and good interpretability. Oyente \cite{oyente} explores execution paths by symbolic execution and uses SMT solving to detect vulnerabilities such as reentrancy and arithmetic bugs. Its main strength is path-level semantic analysis. However, it suffers from path explosion and high analysis cost. SmartCheck \cite{smartcheck} transforms Solidity code into XML format and detects suspicious patterns through XPath rules. This design is simple and efficient. However, its detection ability depends on predefined rules. Slither \cite{slither} performs static analysis by extracting structural information such as abstract syntax and dependency relations. It is practical and widely used. However, it still relies heavily on handcrafted rules and may miss complex semantic vulnerabilities. Securify \cite{security} checks whether contract behaviors satisfy compliance and violation patterns. It provides clear reasoning results. However, its performance depends on the completeness of manually designed security patterns. eThor \cite{eThor} performs sound static analysis for Ethereum smart contracts based on Horn-clause reasoning. It provides stronger theoretical support. However, this comes with higher analysis complexity. VerX \cite{Verx} verifies safety properties through symbolic execution and delayed predicate abstraction. It is suitable for contracts with high security requirements. However, it requires formal property modeling and has high verification cost. EthPloit \cite{EthPloit} combines fuzzing with taint analysis and feedback guidance to generate exploit-oriented test cases. It can better validate triggerable vulnerabilities than pure static analysis. However, its performance is still sensitive to seed quality and path coverage. DFier \cite{DBLP:journals/jnca/WangDLCZ24} verifies suspicious paths by constructing directed inputs for vulnerable locations. This improves validation precision. However, it still depends on the quality of prior suspicious-path localization.

\subsection{\textbf{Deep Learning Method}}

Deep learning has become an important direction for smart contract vulnerability detection because it can learn vulnerability patterns from contract data automatically. Existing studies mainly focus on graph modeling, transfer learning, multimodal fusion, and contrastive representation learning. The method in \cite{tmp} models smart contracts as graphs and performs vulnerability detection through structural representation learning. It shows the value of graph semantics. However, early graph models are still limited in feature richness and robustness design. AME \cite{AME} combines graph neural networks with interpretable graph features and expert patterns. It balances representation learning and interpretability. However, its reliance on expert-designed knowledge may reduce flexibility. DLVA \cite{dlva} extracts control-flow graphs and applies graph convolutional networks for vulnerability analysis. It improves structural modeling ability. However, it mainly focuses on graph topology and does not directly address noisy supervision. ContractWard \cite{8967006} develops automated vulnerability detection models for Ethereum smart contracts in the TNSE context. It shows the effectiveness of learning-based detection. However, robustness to label noise is not its main focus. ContractGNN \cite{10700860} constructs vulnerability subgraphs and uses graph neural networks for Ethereum smart contract vulnerability detection. This design improves vulnerability-oriented representation. However, it still assumes relatively reliable labels during training. Cross-Modality \cite{Cross-Modality} transfers knowledge from source code to bytecode through a teacher--student mutual learning framework. Its main strength is cross-modal knowledge transfer. However, noisy annotations are not explicitly modeled. ESCORT \cite{ESCORT} improves vulnerability detection through deep transfer learning and supports better generalization to unseen vulnerabilities. This is useful in low-resource settings. However, transfer learning alone cannot solve noisy-label effects. EFEVD \cite{EFEVD} enhances feature extraction by clustering contracts and learning graph embeddings for contract communities. It strengthens contract representation at the community level. However, it still relies on standard supervision. CLEAR \cite{clear} introduces contrastive learning to improve the discriminative quality of smart contract representations. It shows the value of contrastive objectives. However, it does not explicitly address robust training under noisy supervision. Vulnsense \cite{Vulnsense} integrates source code, opcodes, graph structures, and language modeling for multimodal vulnerability detection. Its multimodal design enriches semantic information. However, the framework is relatively complex and still depends on label quality.

\subsection{\textbf{LLM-assisted Smart Contract Security Analysis}}

Large language models have recently been introduced into smart contract security research. PSCVFinder \cite{PSCVFinder} applies prompt tuning to smart contract vulnerability detection. This shows that prompt-based methods can support contract analysis. However, the method still depends on the stability of prompt inference. SmartGuard \cite{SmartGuard} enhances vulnerability detection with LLM-assisted analysis. Its main advantage is stronger semantic understanding. However, reliability and hallucination control remain major issues. SCALM \cite{SCALM} uses large language models to identify bad practices in smart contracts. This broadens the scope of contract analysis. However, it is less specialized for precise vulnerability discrimination. Smart-LLaMA-DPO \cite{jaccard} improves explainable smart contract vulnerability detection with a reinforced large language model. Its explainability is attractive. However, stable security judgment still requires stronger task-specific robustness.

\subsection{\textbf{Robust Method}}

Label noise is a practical issue in vulnerability detection because many datasets are automatically labeled by analyzers, scanners, or heuristic rules. SySeVR \cite{SySeVR} shows that syntax-aware and semantic-aware representations can improve vulnerability detection. Its representation design is effective. However, noisy labels are not explicitly handled. Reveal \cite{Reveal} revisits deep-learning-based vulnerability detection and points out that reported performance may not fully reflect practical capability. This highlights the need for more robust evaluation and training. IVDetect \cite{IVDetect} performs vulnerability detection with fine-grained interpretations. It improves feature granularity. However, it does not directly target label-noise correction. ReGVD \cite{ReGVD} studies graph neural networks for vulnerability detection and confirms the importance of graph-based modeling. However, its learning process still assumes conventional supervision. LineVul \cite{linevul} uses a transformer-based model for line-level vulnerability prediction. It improves localization granularity. However, its effectiveness can still be degraded by noisy annotations. Nie et al. \cite{Nie} explicitly investigate label errors in deep-learning-based vulnerability detection and propose confidence learning and differential training strategies. This work clearly shows that label noise is a non-negligible issue in vulnerability datasets. SCL-CVD \cite{SCL-CVD} improves code vulnerability detection by supervised contrastive learning with GraphCodeBERT. It strengthens discriminative representations. However, it is designed for general software rather than smart contracts. VulChecker \cite{vulchecker} uses graph-based vulnerability localization and data augmentation to improve robustness. Its graph design is effective. However, it is still not developed specifically for smart contracts. GCL4SVD \cite{gcl} estimates label-noise transitions through graph confident learning and corrects mislabeled samples. It provides a useful reference for noise-aware training. However, it is still targeted at software vulnerability detection.

\subsection{\textbf{Comparison with Our Method}}

Table~\ref{tab:related_summary} gives a direct comparison between representative studies and our method. Existing program-analysis-based methods have good interpretability, but they usually rely on handcrafted rules or expensive path reasoning. Existing deep-learning-based methods improve representation capability, but most of them assume relatively reliable labels. Existing LLM-based methods improve semantic analysis and explanation, but their reliability is still limited in precise vulnerability detection. Prior robust-learning studies also focus mainly on general software rather than smart contracts. In contrast, our method combines contrastive learning and granular-ball-based robust training for smart contract vulnerability detection. The contrastive part improves the discriminative quality of contract representations, while the granular-ball mechanism improves robustness under noisy supervision. Therefore, our method is designed to achieve both robust representation learning and vulnerability detection in smart   contracts.

\section{Methodology}
\begin{figure*}[t]
    \centering
\includegraphics[width=1\textwidth]{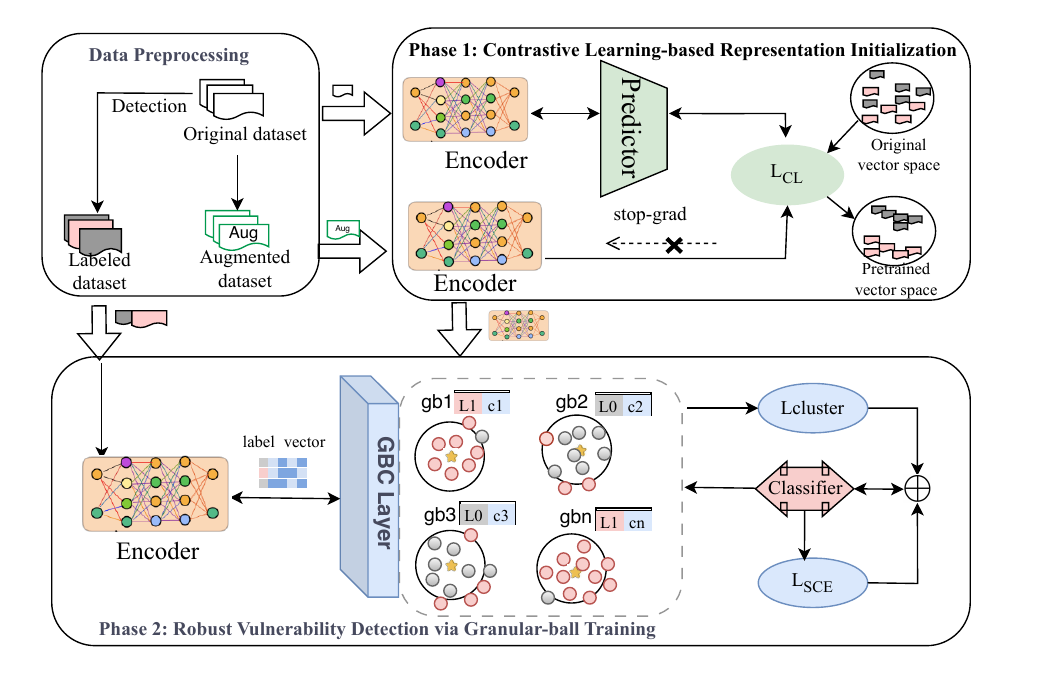}
     \vspace{-0.3cm}
    \caption{The illustration of the overall workflow of CGBC, consisting of two main procedures: contrastive learning-based representation initialization and robust vulnerability detection by Granular-ball training.}
    \label{fig:chart}
    \vspace{-0.4cm}
\end{figure*}

This section illustrates our learning-based vulnerability detection approach for smart contracts (i.e., CGBC), to fight against label noise in training datasets, enhancing the robustness of smart contract vulnerability detection.
CGBC initializes and optimizes the contract encoder based on large-scale unlabeled pertaining, focusing on generating close vector representations for similar smart contracts. Subsequently, CGBC groups smart contracts to create coarse-grained labels and representations during fine-tuning, to correct noisy labels. 
The workflow of CGBC is illustrated in Figure ~\ref{fig:chart}, consisting of two main procedures: (1) contrastive learning-based representation initialization, which introduces a semantic-preserving approach to augment smart contracts, and the extended datasets will pretrain the model without labels based on contrastive learning, to initialize smart contract representation through considering label consistency.
(2) Robust vulnerability detection by Granular-ball training, which instruments a Granular-ball computing layer between the encoder layer and the classifier layer, to generate coarse-grained representations through similarity distance-based clustering, further optimized through a label-resilient symmetric cross-entropy loss function and a two-dimensional clustering loss function.
After training convergence, CGBC derives a robust smart contract vulnerability model despite noisy training datasets.


\subsection{\textbf{Contrastive Learning-based Representation Initialization}}
This module focuses on pretraining the model through unsupervised contrastive learning, based on large-scale positive sample pairs. 
By this procedure, on the one hand, the model can generate more precise vector representations for smart contracts; on the other hand, the representations of similar smart contracts will be closer.
\subsubsection{\textbf{Smart Contract Augmentation}}
The fundamental principle of contrastive learning is constructing positive sample pairs. Conducting data augmentation on smart contracts is an intuitive way to achieve this. However, it is nontrivial because of the complicated business logics and syntax structures of smart contracts. Our core insight is to insert a compilable contract-free code snippet $\mathfrak{s}$ into a random location $Loc$ of the original contract $C$, which can maintain the functionality of $C$ in such a case.  
\begin{figure*}[t]
    \centering
    \includegraphics[width=\textwidth]{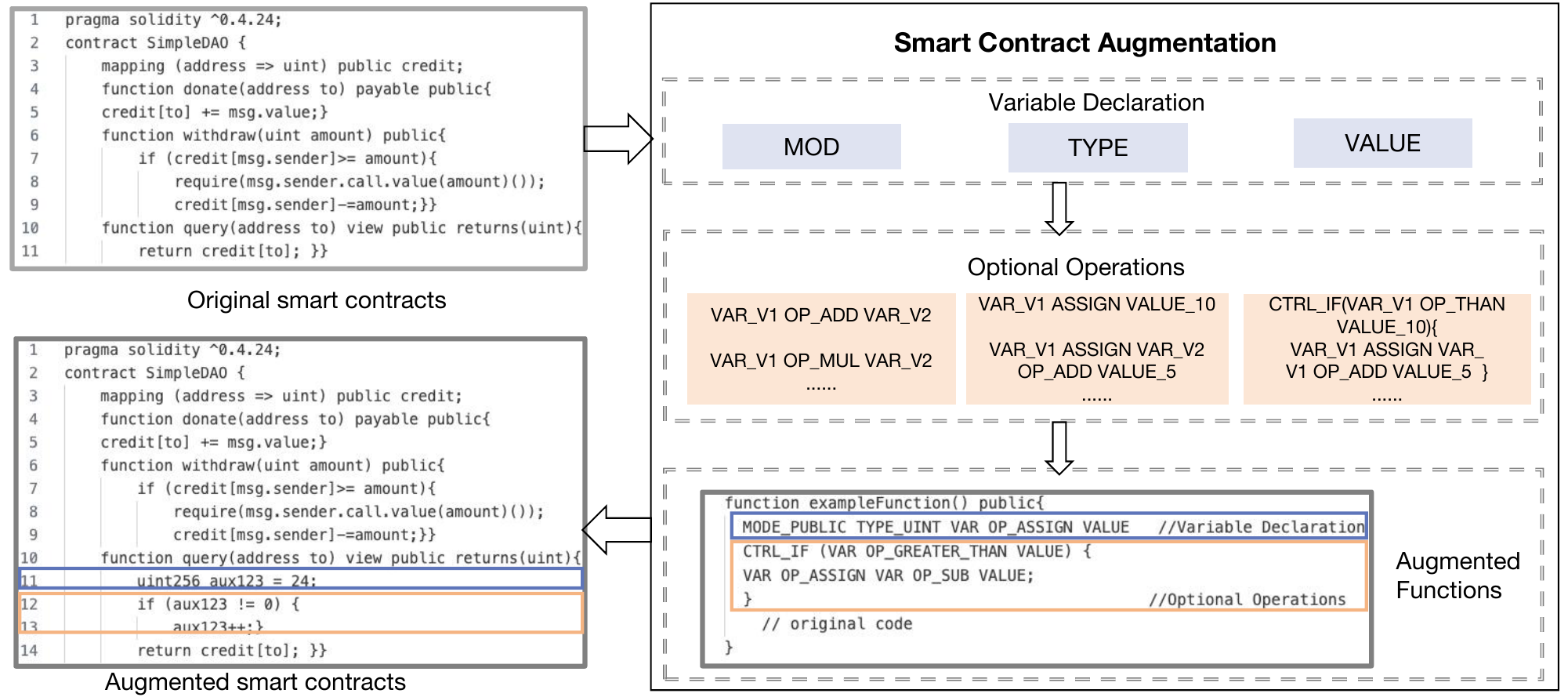} 
    \caption{An illustration example of smart contract augmentation.}
    \label{fig:augmentation_process}
    \vspace{-0.6cm}
\end{figure*}
Specifically, according to the syntax structure of smart contracts, the generation of such a $\mathfrak{s}$ must firstly declare new variables $\mathfrak{s}_1$, then create operation codes $\mathfrak{s}_2$ on these variables to realize contract-free, without interfering with the control and data dependencies of the raw contract $C$. The operations are optional, which will be created after being randomly chosen. We can obtain the augmented code snippet $\mathfrak{s}$ by stacking and combining $\mathfrak{s}_1$ and $\mathfrak{s}_2$. Figure \ref{fig:augmentation_process}
presents the overall contract augmentation process, and Table \ref{tab:formalized_words} describes the meaning of formalized words that will be used later. For declaring variables, the module begins with the random selection of a modifier \( \text{MOD} \) indicating the access attributes such as `public', `private', and `internal', and a variable type \( \text{TYPE} \). Subsequently, a random variable name \( \text{VAR} \) and value \( \text{VALUE} \) are generated. This process is formalized as \( \text{MOD} \ \text{TYPE} \ \text{VAR} \ \text{=} \ \text{VALUE} \). Until now, the generated statements $\mathfrak{s}_1$ can act as a code snippet $\mathfrak{s}$. To further diversify the augmentation samples, CGBC meticulously designs three kinds of operations on the new variables, namely assignment, computation, and condition operations, as the second row of Figure \ref{fig:augmentation_process} shows. The assignment operations \( \text{ASSIGN} \) are employed on the multiple type-identical variables declared in the first snippet $\mathfrak{s}_1$. The computation operations \( \text{COMP} \) involve arithmetic \( \text{ARI} \) and logical \( \text{LOGI} \) operations: arithmetic operations mainly handle numerical data (e.g., integers, floating-point numbers); logical operations focus on Boolean values (True/False) or binary bits (0/1); CGBC will select the appropriate computation method based on the operand type. 
The new variable can also participate in conditional operations \text{CTRL} to further enhance the syntactic complexity and semantic depth of the augmentation snippet, including \texttt{if}, \texttt{while}, and \texttt{for} et al. The generation of control statements is formalized as \( \text{CTRL} \ \text{(} \ \text{VAR} \ \text{OP} \ \text{VALUE} \ \text{)} \ \text{\{} \ \text{VAR} \ \text{OP} \ \text{VAR} \ \text{OP} \ \text{VALUE} \ \text{\}} \). For instance, an \texttt{if} statement checks a condition involving a variable and performs an operation based on whether the condition is true or false. Similarly, \texttt{while} and \texttt{for} loops introduce iterative behavior, allowing the code to execute a block of statements repeatedly under certain conditions. By incorporating these control structures, the augmented code snippets can simulate more realistic and complex behaviors in real-world smart contracts.
Finally, CGBC will organize these statements to create the code snippet $\mathfrak{s}$. Through this rich contract-free code-generating procedure, the snippet $\mathfrak{s}$ can be arbitrarily inserted into any smart contract, to guarantee diversity without influencing its raw functionality. Specifically, for a given contract $C$, we first identify all its functions, denoted as \( \text{FUNC}[0..n] \), where \( n \) represents the number of functions. Subsequently, \( m \) functions are randomly selected from this set as the augmentation targets, denoted as \( \text{SELECTED\_FUNCS} \), with the value of \( m \) dynamically determined based on the size of \( n \). For each selected function \( \text{FUNC} \in \text{SELECTED\_FUNCS} \), an augmented code snippet $\mathfrak{g}$ is generated as mentioned above and inserted into a random location $Loc$ in \(\text{FUNC} \). Iterating this procedure, augmenting \( m \) \( \text{SELECTED\_FUNCS} \) and inserting them into the original contract $C$, we can get an augmented sample $\widetilde{C}$. 

To ensure that augmented contract samples maintain consistent semantics with the original contracts, we design an automated method for comparing semantic similarity. Specifically, for a given pair of original and augmented contracts, we first employ static analysis tools to extract the sequence of core operations at the intermediate representation (IR) level from the target function. These operations encompass state variable accesses (read/write), external function calls, event emissions, return statements, conditional branches, and critical built-in Solidity function invocations. Formally, for a contract file $F$, contract name $C$, and function signature $S$, we denote the target function as $f_{F, C, S}$, and its core operation sequence as
\begin{equation}
O = [o_1, o_2, ..., o_n],
\label{eq:op_sequence}
\end{equation}
where each $o_i$ represents one of the operation types aforementioned. We then define a semantic encoding function $f_{\text{sem}}(\cdot)$ that maps the operation sequence $O$ to a unique semantic digest $D$ via a cryptographic hash:
\begin{equation}
D = f_{\text{sem}}(O) := \mathcal{H}(o_1 \Vert o_2 \Vert \cdots \Vert o_n),
\label{eq:semantic_encoding}
\end{equation}
where $\mathcal{H}(\cdot)$ denotes the SHA256 hash function and $\Vert$ denotes sequence concatenation.
During the detection phase, we generate semantic digests $D_1$ and $D_2$ for the target functions of the original contract $(F_1, C_1, S_1)$ and the augmented contract $(F_2, C_2, S_2)$, respectively. If $(D_1 = D_2)$, the two functions are considered semantically equivalent at the core operation level; otherwise, a semantic deviation is assumed. To balance efficiency and representativeness, we randomly select 1,000 pairs of augmented contract samples for semantic similarity verification. The results demonstrate a semantic similarity rate of 99.60\%, indicating that the generated augmented samples almost entirely preserve the semantics of the original contracts at the core operation level.

Beyond semantic equivalence, we further assess the structural diversity of the augmented samples. We adopt a \emph{Jaccard similarity}-based approach \cite{jaccard} to quantify differences between the original contract and its augmented counterparts, as well as among the augmented samples themselves. The source code is tokenized into identifiers, keywords, operators, numbers, and delimiters, forming a token sequence from which contiguous $k$-grams are extracted. For each pair of code sequences, the Jaccard similarity is computed as the ratio of the intersection over the union of their $k$-gram sets. A similarity threshold $\tau$ is then used to distinguish structurally similar samples from sufficiently diverse ones. In our experiments, with five augmented samples generated per contract, we observe that under a threshold of 0.9, 49.49\% of the comparisons indicate structural diversity, whereas relaxing the threshold to 0.95 increases this proportion to over 74.03\%. These results demonstrate that our augmentation strategy not only preserves semantic correctness but also introduces substantial structural variation, providing a solid foundation for effective contrastive representation learning.


\input{Table/formalized_words}

\subsubsection{\textbf{Contract Representation Pretraining}}
Given a batch of pairs containing a smart contract $\mathcal{C}$ together with its augmentation contract $\widetilde{\mathcal{C}}$, this module aims at dragging the vector representations of similar smart contracts closer, and improving the expressive ability of the DNN model through large-scale unsupervised Contrastive Learning.
Specifically, $\mathcal{C}$ and $\widetilde{\mathcal{C}}$ are fed into a contrastive learning model consisting of two shared-weight encoders $\mathcal{F}_1$ and $\mathcal{F}_2$. The encoders will output their vector representations \(z \in \mathbb{R}^d\)  and \(\widetilde{z} \in \mathbb{R}^d\) as shown in Equation (\ref{equ: embed}). 
\begin{equation}
z = \mathcal{F}_1(\mathcal{C}), \quad 
\widetilde{z} = \mathcal{F}_2(\widetilde{\mathcal{C}})
\label{equ: embed}
\end{equation}
Followed by a predictor $\mathcal{P}$, a two-layer multi-layer perceptron (MLP), the predictor $\mathcal{P}$ performs a nonlinear transformation on one branch of the vector representation \( z\) or \( \widetilde{z} \), generating the predication vector \( p \) or \( \widetilde{p} \) as follows:
\begin{equation}
p = \mathcal{P}(z), \quad \widetilde{p} = \widetilde{\mathcal{P}}(\widetilde{z})
\end{equation}
where \( p, \widetilde{p} \in \mathbb{R}^d \). Then, a symmetric loss function $\mathcal{L}$ is employed to measure and optimize network performance. based on cosine similarity. The first step of $\mathcal{L}$ is to compute the cosine similarity loss between the prediction vector \( p \) and the vector representation \( z \):
\begin{equation}
\mathcal{L}_{\text{cos}}(p, z) = 1 - \frac{\langle p, z \rangle}{\|p\| \|z\|}
\end{equation}
where \( \|.\| \) is the L2 norms, and \( \langle p, z \rangle \) denotes the dot product of the vectors. By adding the constant term 1, the cosine similarity value is mapped to a non-negative value, which is then used for loss optimization. The final symmetric loss function is defined as:
\begin{equation}
\mathcal{L} = \frac{1}{2} \left( \mathcal{L}_{\text{cos}}(p, \widetilde{z}) + \mathcal{L}_{\text{cos}}(\widetilde{p}, z) \right)
\end{equation}
This involves calculating the similarity between the prediction vector \( p \) and the vector representation \( z \), as well as between \( \widetilde{p} \) and \( \widetilde{z} \), and then averaging the results. This design ensures that the loss function treats both branches symmetrically, thus preventing bias during the optimization process. Moreover, by integrating stop-gradient operations into this negative-sample-free paradigm, we maximize the mutual information between augmented contract pairs, thereby achieving discriminative representation learning purely through unsupervised feature alignment.

\subsection{\textbf{Robust Vulnerability Detection by Granular-ball Training}}
This module aims to conduct robust fine-tuning based on annotated smart contracts to endow the model with vulnerability detection ability. To diminish the influence of label noise in annotated datasets, CGBC introduces a Granular-ball computing (GBC) layer between the encoder layer and the classifier layer. The GBC layer can cluster close smart contracts and create coarse-grained representations and labels for training samples, improving robustness. This procedure consists of two main steps, namely contract Granular-ball construction and loss-guided vulnerability detection optimization.

\subsubsection{\textbf{Contract Granular-ball Construction}}
The GBC layer creates and optimizes smart contract clustering to generate robust representations under a noisy environment. 
The complete procedure is detailed in Algorithm~\ref{algo:granular_ball}. The clustering process begins with an initial queue \( Q \) containing the full training dataset $\mathcal{D}$ and initializes an empty Granular-ball set \(\mathcal{B}\) (line 1). $\mathcal{D}$ contains a batch of samples (\textbf{x}, y), where \textbf{x} is the embedding of the corresponding smart contract, and y is its label. For each iteration, a cluster $GB$ is popped from the cluster queue \( Q \). When the number of samples in $GB$  or the quality of $GB$ reaches the corresponding thresholds, $GB$ is regarded as an eligible Granular-ball. The quality is measured by \emph{Purity} computed as shown in Equation (\ref{equ:purity}), where \(\mathbf{1}(y_i = y_{\text{max}})\) is an indicator function and $y_{\text{max}}$ represents the label with most samples in $GB$. That is, the metric \emph{Purity} indicates the proportion of samples with the label $y_{max}$. 
If the purity of $GB$ exceeds the quality threshold \(pur\), it indicates that most samples in $GB$ belong to the same category, and the subset is considered pure.
\begin{equation}
\text{Purity}(X) = \frac{1}{|X|} \sum_{i=1}^{|X|} \mathbf{1}(y_i = y_{\text{max}})
\label{equ:purity}
\end{equation}
For the eligible Granular-ball $GB$, their centers and labels are calculated as Equations (\ref{eq:center}) and (\ref{eq:radius}), where the center is the mean of all samples' embeddings, and a majority vote determines the label. This cluster $GB$, together with its center and label, will be stored in the final outputs (lines 5-7). 
Otherwise, the algorithm identifies the two most distant samples ($x_i, y_i$) and ($x_j, y_j$) in $GB$ based on cosine distance as shown in Equation (\ref{eq:cosinedistance}) ( lines 9-10). Its intuition is that the two most distant samples are likely to belong to different semantic spaces, thus serving as natural anchors for partitioning. These two anchor points are used to partition $GB$ into two smaller groups $GB_i$ and $GB_j$ according to proximity (lines 11). The new clusters will be added to the queue $Q$ that is pending to be processed. Iterating this procedure until the queue is empty, the original dataset $\mathcal{D}$ will be divided into multiple multi-granularity \emph{Purity}-qualified Granular-balls. As described above, the whole procedure of Granular-ball construction is adaptive without set constraints on clusters (such as the size or number of sets), which is more flexible and robust.
\begin{equation}
\mathbf{c} = \frac{1}{|\mathbf{X}_c|} \sum_{x_i \in \mathbf{X}_c} x_i
\label{eq:center}
\end{equation}
\begin{equation}
y_c = \text{MajorityVote}(\mathbf{y}_c)
\label{eq:radius}
\end{equation}
\begin{equation}
\text{CosineDistance}(x_i, x_j) = 1 - \frac{x_i \cdot x_j}{\|x_i\| \|x_j\|}
\label{eq:cosinedistance}
\end{equation}

\input{algorithm/gbClusteringAndTraining}

\subsubsection{\textbf{Loss-guided Vulnerability Detection Optimization}}
Since most samples of a \emph{Purity}-qualified GB are similar with the same label, the center of it represents the universal and effective features of samples in the current GB. Bearing the core insight that noisy labels are inevitable but are the minority in the large-scale dataset, the embedding and label of all samples in a GB are represented by the GB center and label, respectively. This can improve robustness, but its performance on smart contract vulnerability detection is limited to the accuracy of new coarse-grained representations. To address this problem, we elaborately designed our loss function with two main goals: one is optimizing the contract Granular-ball (GB) to facilitate only similar smart contracts in the same GB, while the other is improving the accuracy of the model on the contract vulnerability detection task.  
For the first goal, we design a clustering loss function composed of intra-GB compactness loss (Definition \ref{def: intra-loss}) and inter-GB looseness loss (Definition \ref{def: inter-loss}) as follows. The former is to ensure that samples within each GB are closer to the center, while the latter makes the GB centers with the same label closer and pushes GB with different labels farther apart. This design enhances the accuracy of coarse-grained representation by improving the clustering performance.

\begin{definition} [Intra-GB Compactness Loss]
    Given a set of Granular-balls $[ GB_1, ..., GB_i, ..., GB_k ]$ generated in the current batch, each $GB_i$ contain $N_k$ samples, where the sample feature vectors are denoted as \( x \) and the center of $GB_i$ is \( c_i \), the compactness loss is calculated as:
    \begin{equation}
\mathcal{L}_{\text{intra}} = \sum_{i=1}^{k}(\frac{1}{N_i} \sum_{j=1}^{N_i} \left\| x_j - c_i \right\|_2^2), \quad x_j \in GB_i
\end{equation}
    which is conducted on all Granular-balls generated by the GBC layer in the current batch. For each Granular-ball, the average squared Euclidean distance between the current Granular-ball center and all its contained samples is computed. 
    \label{def: intra-loss}
\end{definition}

\begin{definition} [Inter-GB Looseness Loss]
    Given a set of Granular-balls $[ GB_1, ..., GB_i, ..., GB_k ]$ generated in the current batch, where each $GB_i$ has a center vector \( c_i \) and a corresponding label \( y_i \), the inter-cluster Separation loss is calculated as:
    \begin{equation}
\mathcal{L}_{\text{inter}} = \sum_{i=1}^{k} \sum_{j=i+1}^{k}
\begin{cases}
\left\| c_i - c_j \right\|_2^2, & \text{if } y_i = y_j \\
\max(0, m - \left\| c_i - c_j \right\|_2), & \text{if } y_i \ne y_j
\end{cases}
\end{equation}
    where $m$ is a margin hyperparameter that enforces a minimum distance between Granular-balls with different labels, setting as $1.0$ to provide sufficient separation between different Granular-balls. When $y_i = y_j$, the loss minimizes the squared Euclidean distance between the centers of same-label clusters. When $y_i \ne y_j$, the loss pushes the dissimilar groups apart by a margin.
    \label{def: inter-loss}
\end{definition}





To improve the quality of Granular-ball construction, we define a clustering objective, denoted by $\mathcal{L}_{\text{clustering}}$, by combining the intra-GB compactness loss and the inter-GB separation loss. This objective encourages samples in the same Granular-ball to stay close to their center while keeping different Granular-balls well separated, thus producing tighter and more reliable coarse-grained representations.

For vulnerability detection, we adopt the symmetric cross-entropy loss, denoted by $\mathcal{L}_{\text{SCE}}$ \cite{sl}, as the classification objective. Compared with standard cross-entropy, $\mathcal{L}_{\text{SCE}}$ is more robust to noisy labels because it balances fast optimization and noise tolerance.

The final loss of CGBC jointly considers clustering quality and classification performance:
\begin{equation}
\mathcal{L} = \mathcal{L}_{\text{clustering}} + \mathcal{L}_{\text{SCE}}.
\end{equation}

\section{Experimental  Evaluation}
To comprehensively evaluate the effectiveness and robustness of CGBC, we design a series of experiments aimed at answering the following three research questions:
\begin{enumerate}[0]
    \item[$\bullet$] \textbf{RQ1:} How effective is CGBC in detecting smart contract vulnerabilities compared to state-of-the-art methods?%
     \item[$\bullet$] \textbf{RQ2:} How do the core components of CGBC contribute to its performance?
     \item[$\bullet$] \textbf{RQ3:} How robust is CGBC under different noise levels to detect vulnerabilities in smart contracts?
     \item[$\bullet$] \textbf{RQ4}: How effective is Granular-ball computing for clustering in smart contracts?
\end{enumerate}
\subsection{\textbf{Experimental Setup}}
\stitle{Datasets.}
To comprehensively evaluate the effectiveness of CGBC in detecting smart contract vulnerabilities, we curate training and evaluation datasets as follows: (1) Large-scale and Representative Source: The dataset should originate from publicly accessible, large-scale corpora that represent diverse real-world smart contracts. (2) Verified Vulnerability Labels: For fine-tuning and evaluation, contracts must be annotated with confirmed vulnerability types (e.g., reentrancy, integer overflow/underflow, timestamp dependency). (3) Tool-Detectable Weaknesses: The labeled vulnerabilities should be detectable by mainstream static analysis tools to ensure compatibility and reproducibility. Our experiments leverage three key datasets. For pretraining, we adopt Ethereum-SC Large\cite{dlva}, a comprehensive corpus with over 113,000 raw smart contracts. After preprocessing to remove invalid and duplicate entries, 78,211 contracts remain for self-supervised representation learning. During fine-tuning, we use a labeled dataset from \cite{Cross-Modality}, which contains approximately 40,000 real-world smart contracts annotated with major vulnerability types such as reentrancy, integer overflow/underflow, and timestamp dependency. For evaluation, we aggregate contracts from two widely accepted benchmarks: ReentrancyBenchmark\cite{dlva} and SolidiFiBenchmark\cite{dlva}. These datasets provide verified labels derived from manual annotation and static analysis tools. Each vulnerability type is represented by a binary label per contract, where 0 indicates secure and 1 indicates vulnerable. The vulnerability detection is conducted at the contract level. To elude data leakage across the pretraining, fine-tuning and testing sets, we identify and remove their overlapping samples. Specifically, we normalize source codes and compute their SHA-256 hashes to detect identical contracts. This check proves the isolation of the testing set and identifies 4,781 groups of identical contracts between pre-training and fine-tuning sets; these duplicates were removed from the fine-tuning data to ensure strict separation. Intra-set duplication was also detected, with 1,265 redundant groups in the pre-training set and 204 in the fine-tuning set, all of which were subsequently deleted.

\stitle{Implementation Details.} To ensure the effective extraction of semantic and structural information from smart contracts, we design a contract encoder based on a multi-layer Transformer architecture. The encoder operates on tokenized contract sequences with a maximum length of 512 and is trained with a masking strategy that encourages the model to infer contextual relationships between tokens under partially observable input. This setting strengthens the model’s ability to capture latent code semantics and function-level dependencies, which are essential for vulnerability detection. Following the encoder, we stack several fully connected layers to refine the learned representation and project it into a latent space suitable for downstream classification. The model processes two views of the same input contract in parallel, enforcing representation consistency across syntactically different but semantically equivalent inputs. 
For the effectiveness and ablation experiments, the corresponding methods are first pretrained on the dataset\cite{dlva} and then finetuned on the dataset\cite{Cross-Modality}. The results of Granular-ball clustering is derived during finetuning CGBC.
In addition, the noise-injection experiments are performed by perturbing the fine-tuning dataset, and all evaluations are consistently carried out on the benchmark dataset. All experiments are launched on TensorFlow-GPU 2.8.0 with a learning rate of 1e-5 and run on four Tesla V100-SXM2-32GB GPUs. All experimental results are the average of five independent runs.

\input{Table/effectiveness1}
\subsection{\textbf{Overall Evaluations}}
\stitle{Evaluations on RQ1: Effectiveness in detecting vulnerability.} Table~\ref{tab:comparison} reports the results of 16 vulnerability detection methods on three common smart contract vulnerabilities: Reentrancy (RE), Timestamp Dependency (TD), and Integer Overflow (IO). We use recall (R), precision (P), and F1-score (F) as evaluation metrics. All baseline results are taken from \cite{clear,TMF-Net}. For fairness, CGBC is trained and tested on the same datasets. Some tools, such as Osiris, Oyente, and Securify, do not support all vulnerability types, so their results are only reported on the tasks they cover.

We first compare CGBC with seven rule-based static analysis tools. These methods mainly rely on handcrafted rules, so they often struggle with complex contract logic. CGBC clearly outperforms them on all supported tasks. For RE, CGBC reaches an F1-score of 94.74\%, while Slither obtains 73.97\%. For TD, CGBC achieves 97.14\%, much higher than SmartCheck’s 59.73\%. For IO, CGBC reaches 98.36\%, far above Mythril’s 66.80\%. These results show that CGBC can capture vulnerability patterns more effectively than rule-based tools.

Next, we compare CGBC with five advanced deep learning methods. Although these methods learn better features than static tools, they are still affected by label noise and limited contract semantics. Among them, DMT gives the best average F1-score, which is 86.14\%. CGBC achieves 96.75\%, improving the average F1-score by 10.61\%. The gain on IO is especially large, reaching 17.23\%. This result suggests that the coarse-grained Granular-ball representation helps the model learn more robust features. We also compare CGBC with LineVul, a general pretrained code model. Its average F1-score is 76.57\%, which shows that directly using a general code model is not enough for smart contract vulnerability detection. CLEAR achieves an average F1-score of 94.52\%, establishing a strong baseline. As a more recent method, TMF-Net reports an average F1-score of 86.23\%, outperforming most previous deep learning baselines. Even so, CGBC still performs better than both CLEAR and TMF-Net, and achieves the best average F1-score of 96.75\%, with the highest F1-scores on TD and IO, 97.14\% and 98.36\%, respectively.

\begin{answerbox}
\textbf{Answer to RQ1:} CGBC consistently outperforms traditional static analysis tools, general pretrained code models, and recent state-of-the-art deep learning methods, achieving the best average precision and F1-score of 95.99\% and 96.75\%. This demonstrates its strong effectiveness and robustness in detecting diverse smart contract vulnerabilities.
\end{answerbox}

\begin{figure*}[t]
  \centering
  \subfloat[Results of the precision (\%) ]{%
    \includegraphics[width=0.32\linewidth]{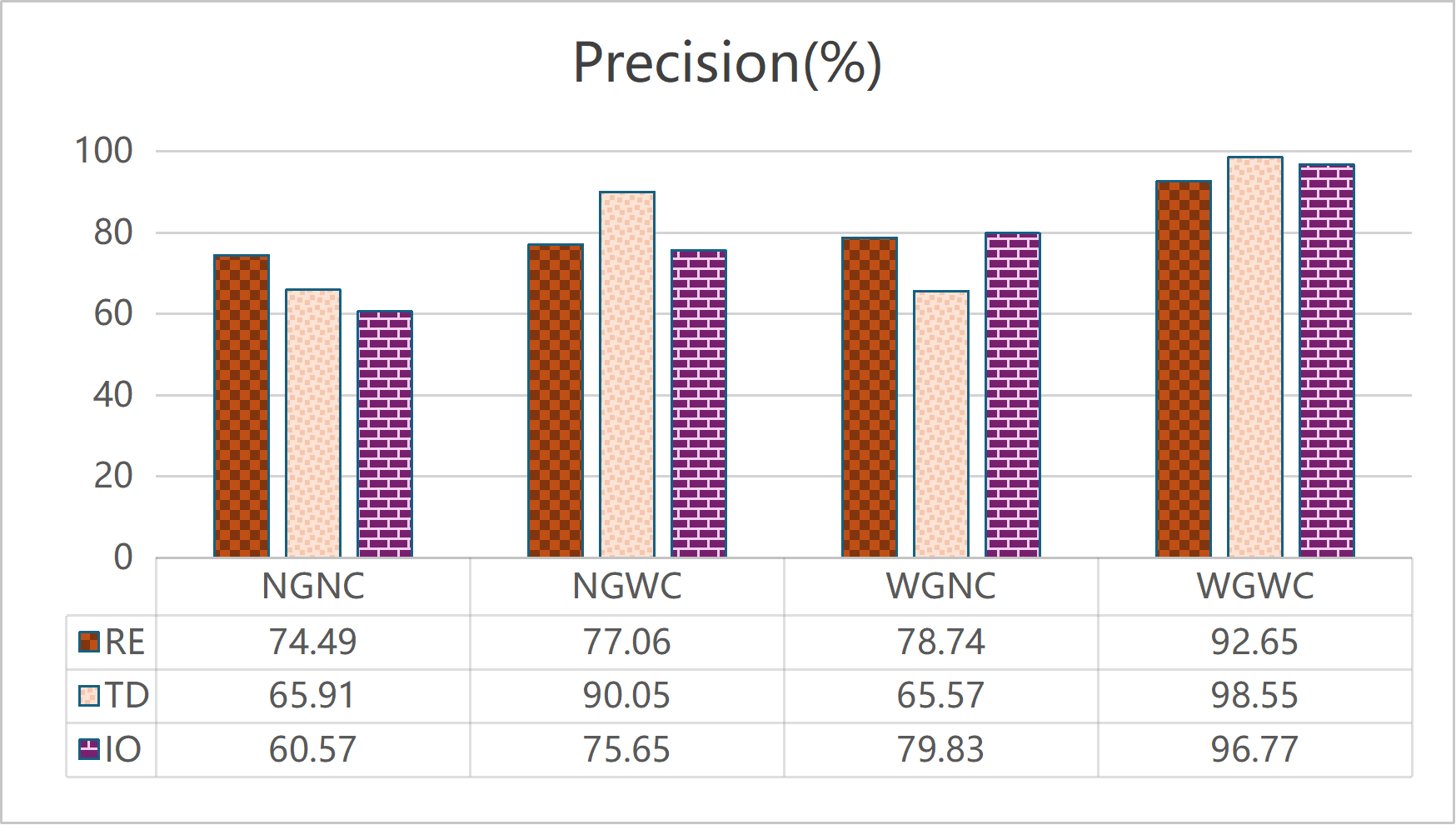}%
    \label{subfig:precision}
  }
  \hfill
  \subfloat[Results of the recall rate (\%)]{%
    \includegraphics[width=0.32\linewidth]{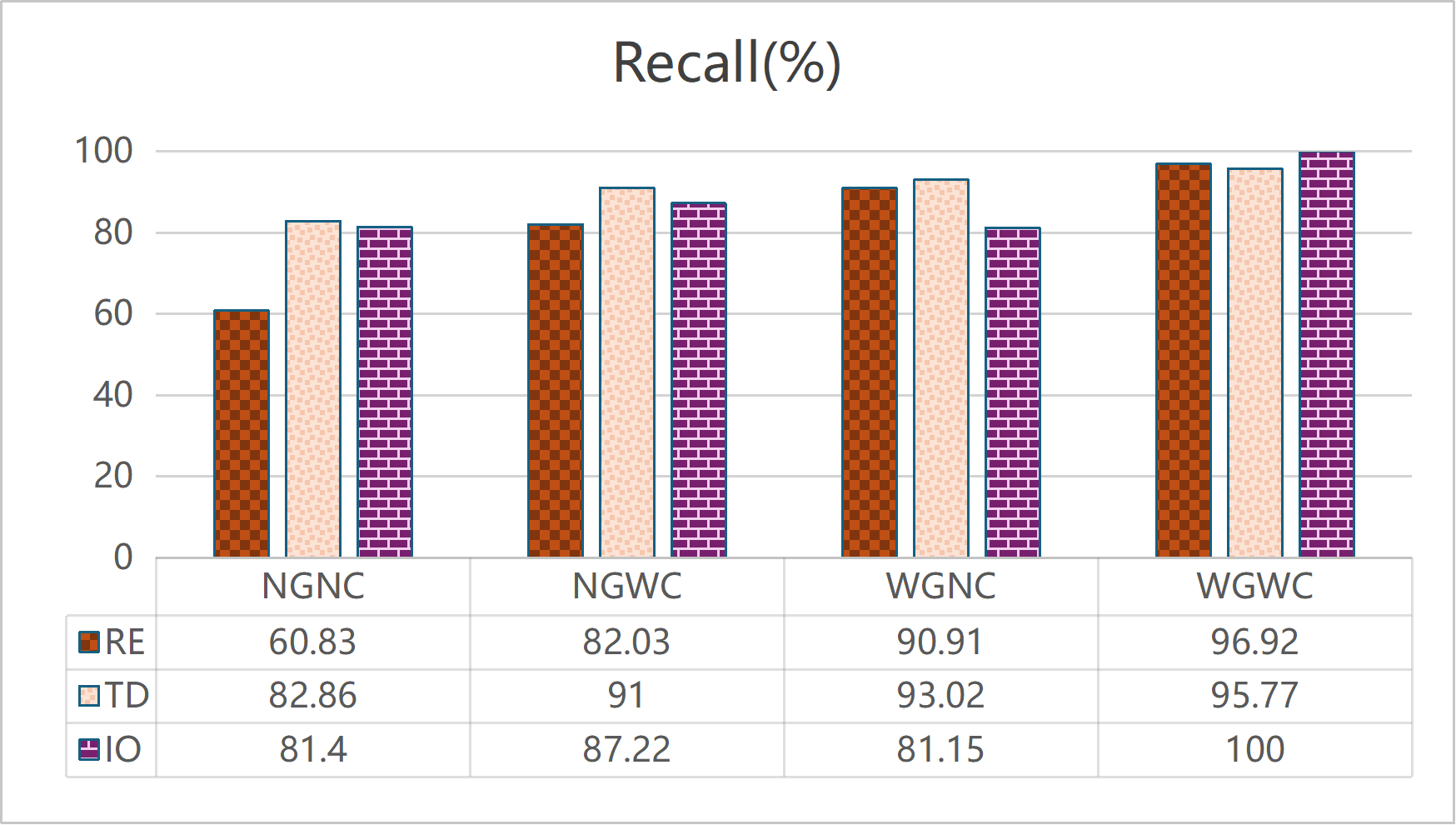}%
    \label{subfig:recall}
  }
  \hfill
  \subfloat[Results of the F1-score (\%)]{%
    \includegraphics[width=0.32\linewidth]{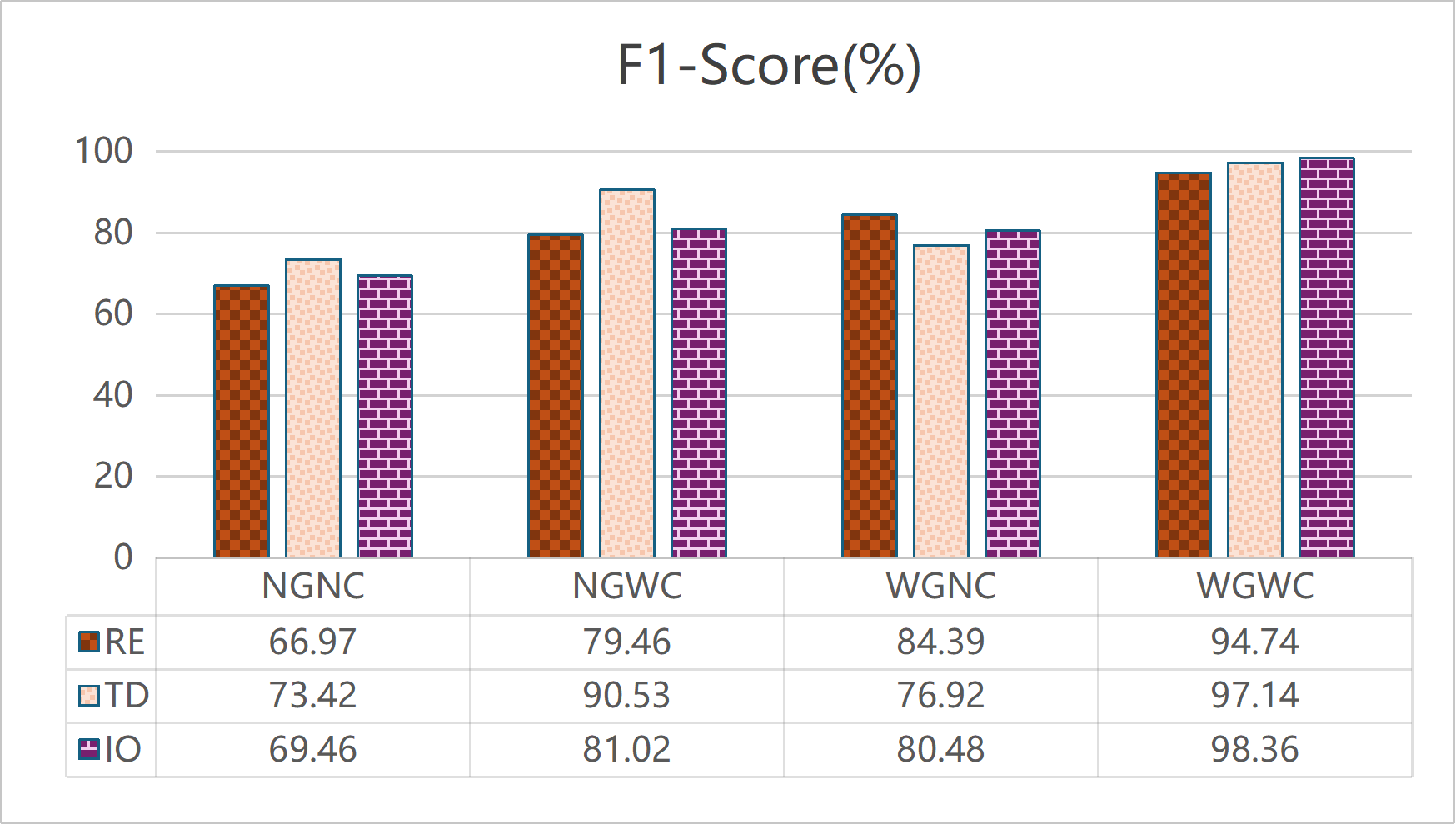}%
    \label{subfig:f1}
  }
  \caption{Evaluation results on RQ2: ablation studies on the core modules of CGBC.}
  \label{fig:ablation}
\end{figure*}

\stitle{Evaluations on RQ2: Ablation Studies.}
The overall trend is clear. Models with CL or GB perform better than the plain model NGNC, and the model with both modules, WGWC, gives the best results on all three tasks. This shows that both modules are useful, and they work best when used together.
The CL module mainly improves feature learning. After adding CL, the model achieves clear gains on all three tasks. The F1-score increases from 66.97\% to 79.46\% on RE, from 73.42\% to 90.53\% on TD, and from 69.46\% to 81.02\% on IO. This suggests that large-scale pretraining helps the model learn better semantic information from smart contracts.The GB module mainly improves robustness during fine-tuning. After adding GB, the model also achieves better results in most cases. The F1-score rises to 84.39\% on RE, 76.92\% on TD, and 80.48\% on IO. These gains show that coarse-grained representation learning helps reduce the effect of noisy labels.
When CL and GB are used together, the model reaches the best performance on every task. The F1-scores are 94.74\% on RE, 97.14\% on TD, and 98.36\% on IO. These results indicate that CL and GB are complementary. CL improves representation quality, while GB makes the model more robust.

\begin{answerbox}
\textbf{Answer to RQ2:} Both the pretraining module and the Granular-ball training module significantly boost CGBC's performance. Their combination (WGWC) yields the highest scores with an average precision of 95.99\%, recall of 97.56\%, and F1-score of 96.75\%, underscoring their complementary roles and effectiveness in smart contract vulnerability detection.
\end{answerbox}

\vspace{-0.2cm}
\input{Table/noisy}
\stitle{Evaluations on RQ3: Robustness against Label Noise.} To comprehensively assess the robustness of CGBC under label noise, we injected synthetic label noise into the fine-tuning training set at levels of 10\%, 20\%, 30\%, and 40\%. At each noise level, a corresponding proportion of training samples were randomly selected and assigned flipped labels, while the test sets remained clean. This configuration emulates realistic annotation errors and allows a rigorous evaluation of model stability under varying degrees of label corruption. For clarity, we compare CGBC against a baseline model without pretraining and without the Granular-ball layer. Results are reported in terms of precision (P), recall (R), and F1-score (F), with detailed numbers shown in Table~\ref{tab:noisy}.

On the original dataset without added noise, CGBC already outperforms the baseline on all three tasks. Its F1-scores are 94.74\% on RE, 97.14\% on TD, and 98.36\% on IO. The gains over the baseline are 27.77\%, 23.72\%, and 28.90\%, respectively. This suggests that even the original data may contain some noisy labels, and robust training is useful even in this case.
As the noise level increases, the baseline degrades quickly. At 40\% noise, its F1-scores drop to 32.97\% on RE, 27.05\% on TD, and 61.78\% on IO. In contrast, CGBC remains much more stable. At the same noise level, its F1-scores are still 62.50\% on RE, 62.81\% on TD, and 73.44\% on IO. Although CGBC also declines as noise grows, the drop is much smaller than that of the baseline.
This robustness comes from the combination of contrastive pretraining and the GBC layer. The GBC layer reduces the effect of noisy labels by using coarse-grained grouping, while contrastive pretraining helps the model learn better semantic representations.
We also compare CGBC with CLEAR \cite{clear}, which is the closest method to ours. Both methods use a two-stage framework, but CGBC adds coarse-grained fine-tuning for robustness. Table~\ref{tab:clearComparison} shows the comparison under different noise levels. Both methods perform well on clean data, but CGBC is more stable as noise increases. At 40\% noise, CGBC outperforms CLEAR on all three vulnerability types. This further shows the advantage of CGBC in noisy settings.
\begin{answerbox}
\textbf{Answer to RQ3:} CGBC exhibits strong robustness to label noise. Across 30 experiments across various noise levels and vulnerability types, CGBC always outperforms the baseline model. Notably, under severe label noise (40\%), CGBC maintains stable and high F1-scores—62.5\% (RE), 62.81\% (TD), and 73.44\% (IO)—outperforming competitive methods. The results confirm that CGBC delivers superior detection performance in noisy real-world scenarios.
\end{answerbox}

\input{Table/clearCompare}
\begin{figure*}[t]
  \centering
  \subfloat[The early stage of training]{%
    \includegraphics[width=0.3\linewidth]{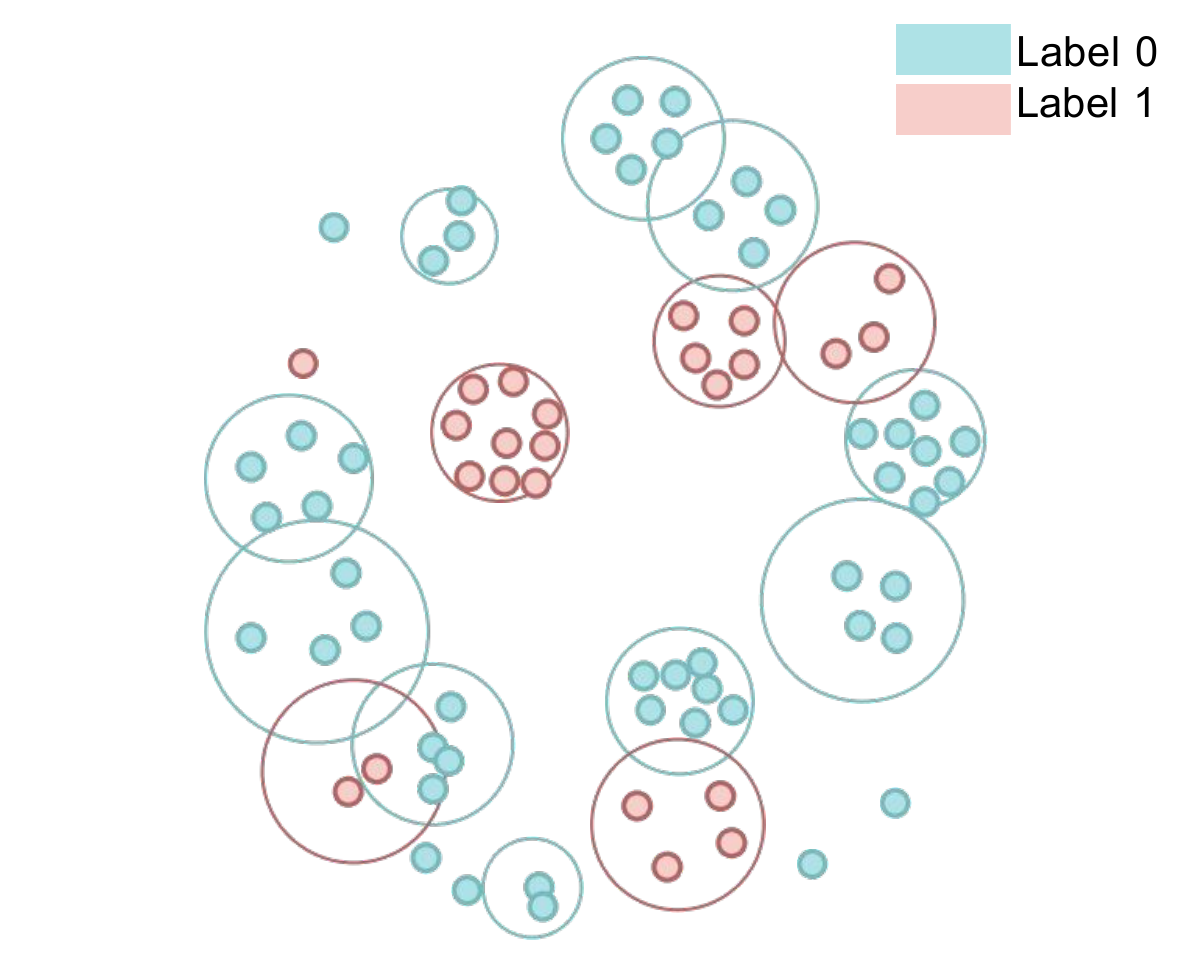}%
    \label{subfig:iniGB}
  }
  \hfill
  \subfloat[The middle stage of training]{%
    \includegraphics[width=0.3\linewidth]{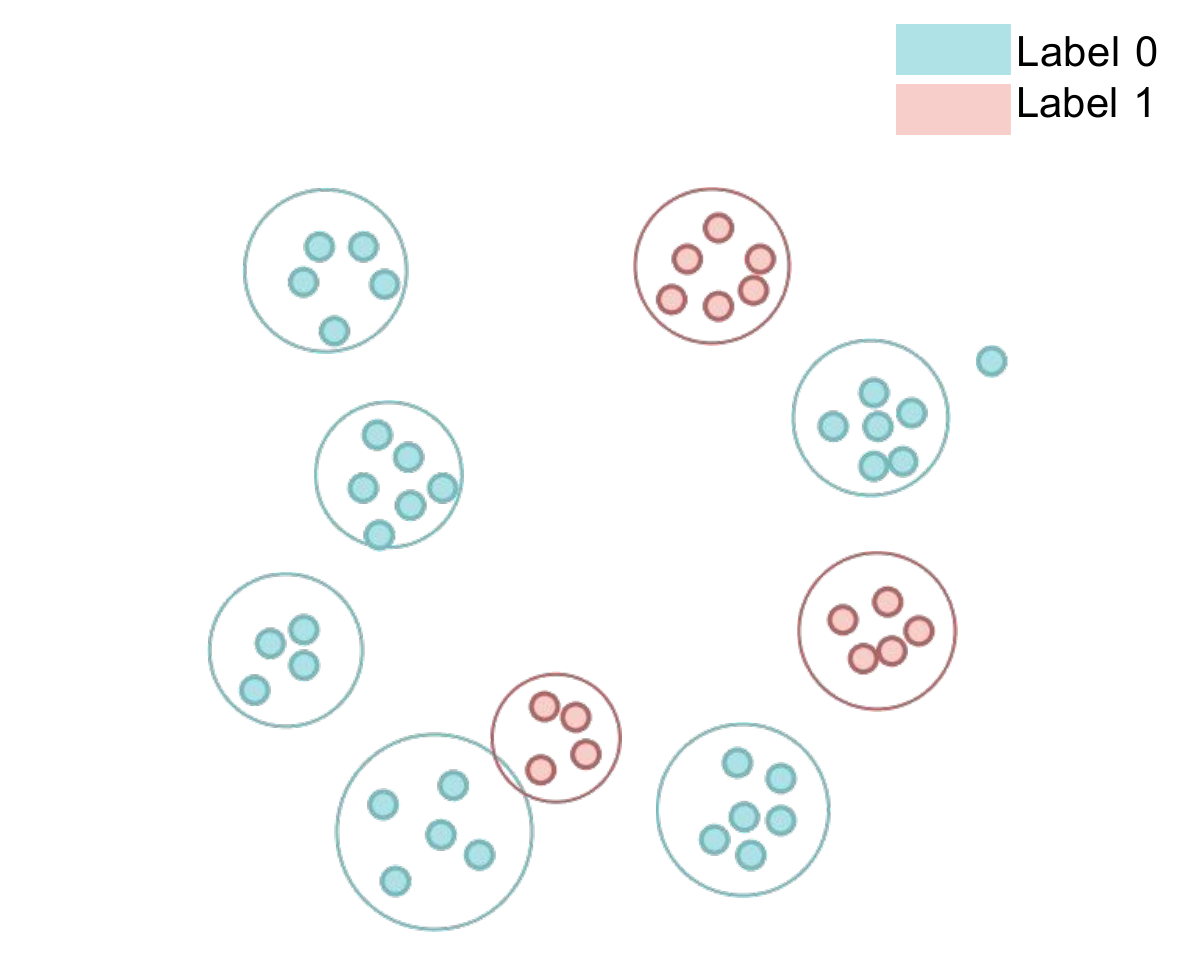}%
    \label{subfig:intermiGB}
  }
  \hfill
  \subfloat[The convergence stage of training]{%
    \includegraphics[width=0.33\linewidth]{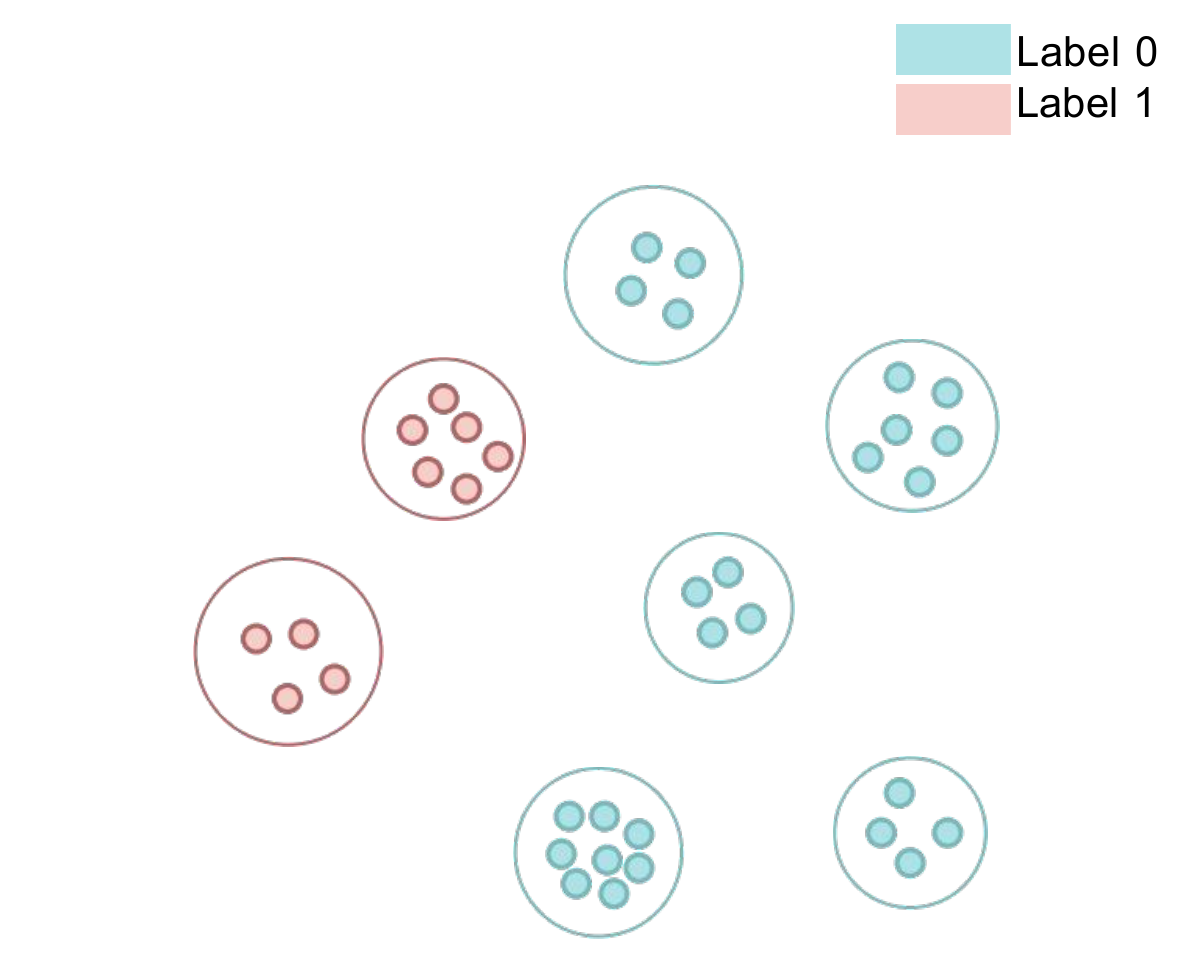}%
    \label{subfig:lastGB}
  }
  \caption{Evaluation results on RQ4: effectiveness of Graunlar-ball computing in clustering smart contracts across training stages.}
  \label{fig:gbcluster}
\vspace{-0.4cm}
\end{figure*}

\stitle{Evaluations on RQ4: Effectiveness of Granular-ball Computing in Clustering Smart Contracts.} To intuitively demonstrate the advantages of Granular-ball multi-granularity and adaptive characteristics, we visualize smart contract clustering results of the Granular-ball layer in different phases of fine-tuning training, as illustrated in Figure \ref{fig:gbcluster}.
The early-stage clustering performance of the GBC layer is demonstrated in Figure \ref{subfig:iniGB}. The smart contract representations are dispersed without identifiable clustering patterns. The formed Granular-balls (GBs) exhibit loose structures with ambiguous boundaries, indicating the GB layer's limited effectiveness at this stage. However, with the help of contrastive learning-based pretraining, the representations of smart contracts with the same label remain relatively close.
As training progresses to the mid stage (Figure \ref{subfig:intermiGB}), the contracts' embeddings within each GB become more compact and the boundaries between the GBs start to emerge more clearly. 
This is because of the feedback and optimization of the intra-GB compactness and inter-GB looseness loss functions, which pull contracts with the same labels closer and push contracts with different labels farther.
When training converges (Figure \ref{subfig:lastGB}), the clustering structure becomes significantly clearer. Smart contracts are densely grouped within their corresponding GBs, and
GBs exhibit well-separated boundaries with distinct margins. This reflects the effectiveness of the GB layer in clustering smart contracts, hinting at its accuracy in coarse-grained representation learning.

\begin{answerbox}
\textbf{Answer to RQ4:} 
The Granular-ball computing layer demonstrates effective clustering performance for smart contracts. During training, the quality of clustering progressively improves. Upon convergence, smart contracts are densely aggregated within subordinate GBs, while the GBs themselves maintain boundaries with distinct margins.
\end{answerbox}

\subsection{\textbf{Threats to Validity}}
Our evaluations face several threats to validity. For \emph{threats to external validity}, they mainly stem from the evaluated datasets and vulnerabilities. To combat such threats, we conduct experiments on several large publicly available real-world datasets. Moreover, we select three popular vulnerability types to evaluate model performance.
For \emph{threats to internal validity} from the implementation of CGBC and the compared approaches. To mitigate these threats, we leverage the widespread SimSiam network framework \cite{simsiam} to implement the contrastive learning module. It is worth noting that the GBC layer inherently involves randomness due to the adaptiveness of GB clustering, which may cause fluctuations in training outcomes. To alleviate their impacts, we performed multiple repeated training runs and reported the average results. The results of baselines are directly derived from work \cite{Cross-Modality}, and we evaluate CGBC on the same datasets when comparing them.

\section{Conclusion}
In this paper, we propose a robust learning-based vulnerability detection method (CGBC) for smart contracts to fight against label noise in datasets. 
Through contrastive learning-based representation pretraining, CGBC can generate discriminative embeddings for smart contracts with different labels. Following this, the Granular-ball computing Layer can adaptively cluster smart contracts and create coarse-grained representations for contracts, diminishing the influence of noisy samples during training.
Extensive experiments demonstrate that CGBC outperforms baseline models and state-of-the-art models across large-scale real-world datasets and various noisy environments. Abundant ablation studies verify the effectiveness of the core modules of CGBC.

\section{ACKNOWLEDGMENTS}
This work was supported by the National Natural Science Foundation of China under Grant Nos. 62402399, 62221005, 62450043, 62222601, and 62176033, Natural Science Foundation of Chongqing, China under Grant No. CSTB2025NSCQ-GPX1268, and the Science and Technology 
Research Program of Chongqing Municipal Education Commission under Grant No. KJQN202500647.

\bibliographystyle{IEEEtran}
\bibliography{main}
\begin{IEEEbiography}
[{\includegraphics[width=1in,height=1.25in,clip,keepaspectratio]{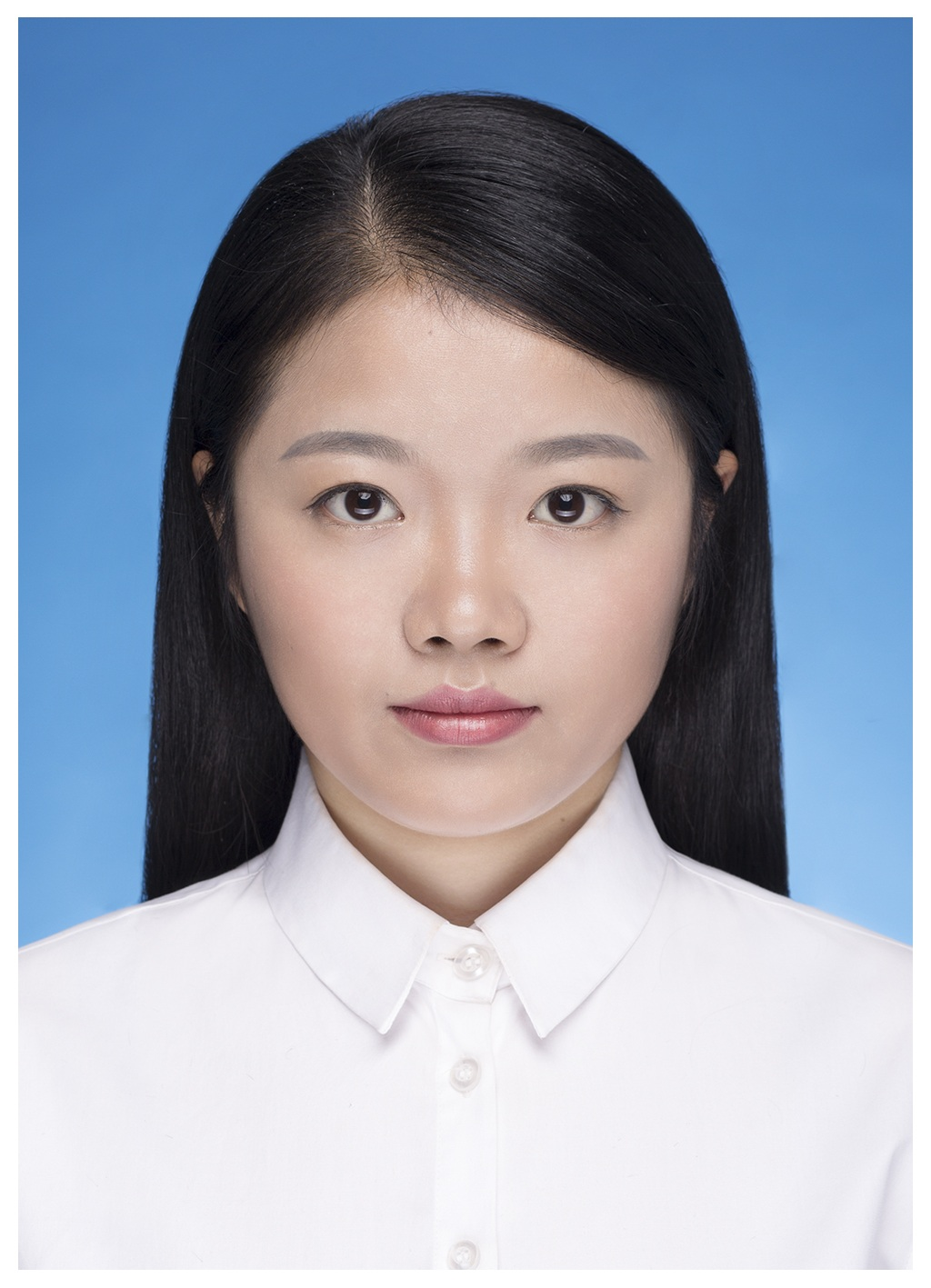}}]{Zeli Wang} received the PhD degree from Huazhong University of Science and Technology (HUST), in 2022. She is now a lecturer at the College of Computer Science and Technology, Chongqing University of Posts and Telecommunications, Chongqing, China. Her current research interests include blockchain smart contract security and the security and robustness of artificial intelligence.
\end{IEEEbiography}

\begin{IEEEbiography}[{\includegraphics[width=1in,height=1.25in,clip,keepaspectratio]{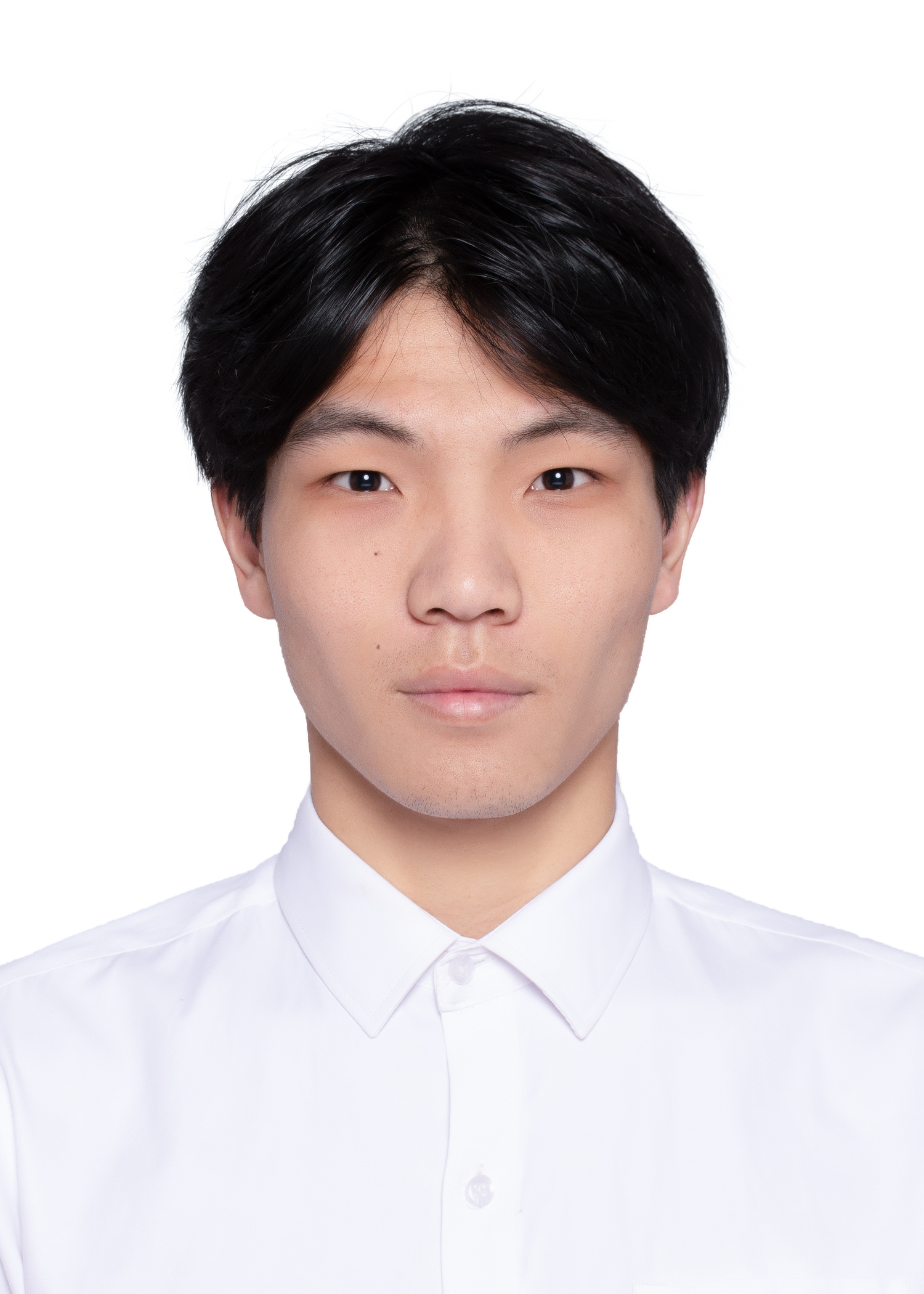}}]{Qingxuan Yang} is currently pursuing an M.E. degree at Chongqing University of Posts and Telecommunications. His main research direction focuses on smart contract security and deep learning.
\end{IEEEbiography}

\begin{IEEEbiography}[{\includegraphics[width=1in,height=1.25in,clip,keepaspectratio]{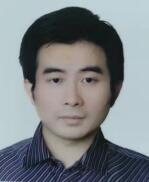}}]{Shuyin Xia} received his B.S. degree and M.S. degree in computer science from Chongqing University of Technology. He received his Ph.D. degree from Chongqing University. He is currently a professor at the School of Artificial Intelligence, Chongqing University of Posts and Telecommunications. He is also the executive deputy director of the Big Data and Network Security Joint Lab of CQUPT. His research interests include classifiers and granular computing.
\end{IEEEbiography}

\begin{IEEEbiography}[{\includegraphics[width=1in,height=1.25in,clip,keepaspectratio]{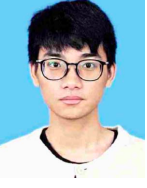}}]{Yueming Wu} received the PhD degree in the School of Cyber Science and Engineering from the Huazhong University of Science and Technology,
Wuhan, China, in 2021. He is currently an associate professor in the School of Cyber Science and Engineering, Huazhong University of Science and Technology, Wuhan, China. His primary research interests lie in malware analysis and vulnerability analysis.
\end{IEEEbiography}

\begin{IEEEbiography}[{\includegraphics[width=1in,height=1.25in,clip,keepaspectratio]{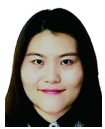}}]{Bo Liu} received the M.S. degree in computer science and engineering from Zhengzhou University and the Ph.D. degree in computer architecture from the Huazhong University of Science and Technology. She is currently a Lecturer with Zhengzhou University. Her research interests focus on deep learning, AI security, and AI accelerator design. She has served as a Reviewer for top conferences or journals, such as IEEE TRANSACTIONS ON PARALLEL AND DISTRIBUTED SYSTEMS and ACM Transactions on Architecture and Code Optimization.
\end{IEEEbiography}

\begin{IEEEbiography}[{\includegraphics[width=1in,height=1.25in,clip,keepaspectratio]{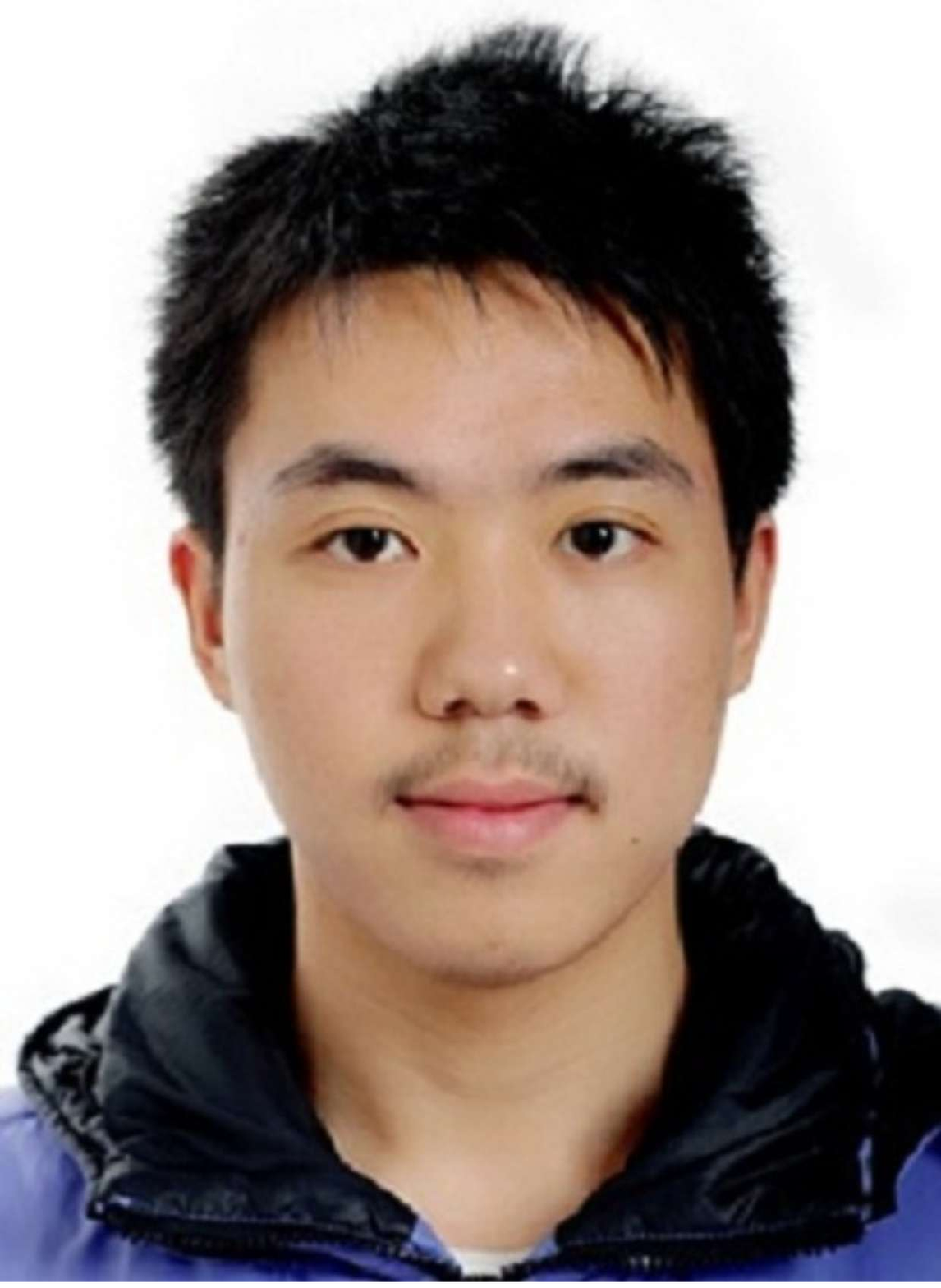}}]{Longlong Lin} received his Ph.D. degree from Huazhong University of Science and Technology (HUST), Wuhan, in 2022. He is currently an associate professor in the College of
	Computer and Information Science, Southwest University, Chongqing. His current research interests include graph clustering and graph-based machine learning.
\end{IEEEbiography}

\end{document}

%% file: Table/formalized_words.tex
\begin{table}[t]
\centering
\caption{Actual meanings of the formalized words in our methods.}

\begin{tabular}{|c|c|c|}

\hline
\textbf{Formalized Word} & \textbf{Meaning} & \textbf{Example} \\ \hline
TYPE & Variable type & `uint256', `bool' \\ \hline
MOD & Modifier & `public', `private' \\ \hline
VAR & Variable name & `V1', `V2' \\ \hline
OP & Arithmetic Operation & `=', `+', `-', `*', `/' \\ \hline
CTRL & Control statement & `if', `while', `for' \\ \hline
\end{tabular}
\vspace{-0.3cm}
\label{tab:formalized_words}
\end{table}

%% file: algorithm/gbClusteringAndTraining.tex
\begin{algorithm}[t]
\caption{Granular-ball Clustering}
\label{algo:granular_ball}
\KwIn{Annotated dataset $\mathcal{D}$, purity threshold $pur$}
\KwOut{Granular-balls $\mathcal{B}$}

 $\mathcal{Q} \gets \mathcal{D}$;$\mathcal{B} \gets \emptyset$;\\
\While{$\mathcal{Q}$ is not empty}{
    $GB = Q.pop()$ \;
    \If{$|GB| \leq 2 \lor \text{Purity}(GB) > pur$}{
        Compute center $\mathbf{x_c} \gets \frac{1}{|\mathbf{X}_c|} \sum_{\mathbf{x} \in GB} \mathbf{x}$ \;
        Compute label $y_c \gets \text{MajorityVote}(\sum_{y \in GB} \mathbf{y})$ \;
        Store $(\mathbf{x_c}, y_c, GB)$ in $\mathcal{B}$ \;
        \textbf{Continue} \;
    }
    Compute pairwise distance matrix $\mathbf{G}$ for $GB$ \;
    $(<x_i,y_i>, <x_j,y_j>) \gets \argmax_{dis(x_i, x_j)} \mathbf{G}$ \;
    $GB_i, GB_j = divide(GB, x_i, x_j)$;
    $Q = Q.add(GB_i, GB_j)$;
    }
    \Return $\mathcal{B}$;
\vspace{-0.1cm}
\end{algorithm}

%% file: Table/effectiveness1.tex
\begin{table*}[t]
\caption{Evaluation results on RQ1: effectiveness on vulnerability detection. The best results of a certain metric for each vulnerability type are highlighted in \textbf{bold}.}
\label{tab:comparison}
\centering
\small
\setlength{\tabcolsep}{4pt}
\renewcommand{\arraystretch}{1.2}
 \scalebox{1.1}{\begin{tabular}{|c|l|ccc|ccc|ccc|ccc|}
\hline
\multirow{2}{*}{\textbf{Line \#}} & \multirow{2}{*}{\textbf{Methods}} 
& \multicolumn{3}{c|}{\textbf{RE}} 
& \multicolumn{3}{c|}{\textbf{TD}} 
& \multicolumn{3}{c|}{\textbf{IO}} 
& \multicolumn{3}{c|}{\textbf{Average}} \\
\cline{3-14}
& & R(\%) & P(\%) & F(\%) 
  & R(\%) & P(\%) & F(\%) 
  & R(\%) & P(\%) & F(\%) 
  & R(\%) & P(\%) & F(\%) \\
\hline
1 & sFuzz\cite{sfuzz}        & 14.95 & 10.88 & 12.59 & 27.01 & 23.15 & 24.93 & 47.22 & 58.62 & 52.31 & 29.73 & 30.88 & 29.94 \\
2 & Smartcheck\cite{smartcheck}            & 16.34 & 45.71 & 24.07 & 79.34 & 47.89 & 59.73 & 56.21 & 45.56 & 50.33 & 50.63 & 46.39 & 44.71 \\
3 & Osiris\cite{osiris}                & 63.88 & 40.94 & 49.90 & 55.42 & 59.26 & 57.28 & n/a   & n/a   & n/a   & n/a   & n/a   & n/a   \\
4 & Oyente\cite{oyente}                & 63.02 & 46.56 & 53.55 & 59.97 & 61.04 & 59.47 & n/a   & n/a   & n/a   & n/a   & n/a   & n/a   \\
5 & Mythril\cite{mythril}               & 75.51 & 42.86 & 54.68 & 49.80 & 57.50 & 53.37 & 62.07 & 72.30 & 66.80 & 62.46 & 57.55 & 58.28 \\
6 & Securify\cite{security}              & 73.06 & 68.40 & 70.41 & n/a   & n/a   & n/a   & n/a   & n/a   & n/a   & n/a   & n/a   & n/a   \\
7 & Slither\cite{slither}               & 73.50 & 74.44 & 73.97 & 67.17 & 69.27 & 68.20 & 52.27 & 70.12 & 59.89 & 64.31 & 71.28 & 67.35 \\
\hline
8 & GCN\cite{gcn}                   & 73.18 & 74.47 & 73.82 & 77.55 & 74.93 & 76.22 & 69.74 & 69.01 & 69.37 & 73.49 & 72.80 & 73.14 \\
9 & TMP\cite{tmp}                   & 75.30 & 76.04 & 75.67 & 76.09 & 78.68 & 77.36 & 70.37 & 68.18 & 69.26 & 73.92 & 74.30 & 74.10 \\
10 & AME\cite{AME}                  & 78.45 & 79.62 & 79.03 & 80.26 & 81.42 & 80.84 & 69.40 & 70.25 & 69.82 & 76.04 & 77.10 & 76.56 \\
11 & SMS\cite{Cross-Modality}                  & 77.48 & 79.46 & 78.46 & 91.09 & 89.15 & 90.11 & 73.69 & 76.97 & 75.29 & 80.75 & 81.86 & 81.29 \\
12 & DMT\cite{Cross-Modality}                  & 81.06 & 83.62 & 82.32 & 96.39 & 93.60 & 94.97 & 77.93 & 84.61 & 81.13 & 85.13 & 87.28 & 86.14 \\
\hline
13 & LineVul\cite{linevul}     & 73.01 & 85.19 & 78.63 & 67.46 & 89.47 & 76.92 & 74.20 & 74.10 & 74.15 & 71.56 & 82.92 & 76.57 \\
\hline
14 & Clear\cite{clear}       & \underline{96.43} & \textbf{96.81} & \textbf{96.62} & \textbf{98.41} & \underline{94.30} & \underline{96.31} & \underline{91.48} & \underline{89.81} & \underline{90.64} & \underline{95.44} & \underline{93.64} & \underline{94.52} \\
15 & TMF-Net\cite{TMF-Net}       & 91.45 & 84.92 & 88.06 & 88.15 & 86.59 & 87.36 & 81.40 & 85.24 & 83.28 & 87 & 85.58 & 86.23 \\
\hline
16 & \textbf{CGBC}       & \textbf{96.92} & \underline{92.65} & \underline{94.74} & \underline{95.77} & \textbf{98.55} & \textbf{97.14} & \textbf{100} & \textbf{96.77} & \textbf{98.36} & \textbf{97.56} & \textbf{95.99} & \textbf{96.75} \\
\hline
\end{tabular}}
\end{table*}

%% file: Table/noisy.tex
\begin{table*}[t]
    \centering
    \small
    \caption{Evaluation results on RQ3: robustness against label noise.}
    \label{tab:noisy}
    \begin{tabularx}{\textwidth}{l *{18}{>{\centering\arraybackslash}X}}
        \toprule
        & \multicolumn{6}{c}{RE} & \multicolumn{6}{c}{TD} & \multicolumn{6}{c}{IO} \\
        \cmidrule(lr){2-7} \cmidrule(lr){8-13} \cmidrule(lr){14-19}
       Noise & \multicolumn{3}{c}{BaseLine} & \multicolumn{3}{c}{CGBC} 
        & \multicolumn{3}{c}{BaseLine} & \multicolumn{3}{c}{CGBC} 
        & \multicolumn{3}{c}{BaseLine} & \multicolumn{3}{c}{CGBC} \\
        \cmidrule(lr){2-4} \cmidrule(lr){5-7} \cmidrule(lr){8-10} \cmidrule(lr){11-13} \cmidrule(lr){14-16} \cmidrule(lr){17-19}
       Levels & P(\%) & R(\%) & F(\%) & P(\%) & R(\%) & F(\%) & P(\%) & R(\%) & F(\%) & P(\%) & R(\%) & F(\%) & P(\%) & R(\%) & F(\%) & P(\%) & R(\%) & F(\%) \\
        \midrule
        0\%  & 74.49 & 60.83 &  66.97 & 92.65 & 96.92 &  94.74 & 65.91 & 82.86 &  73.42 & 98.55 & 95.77 &  97.14 & 60.57 & 81.40 &  69.46 & 96.77 & 100 &  98.36 \\
        10\% & 58.08 & 45.65 &  51.12 & 90.62 & 87.88 &  89.23 & 76.11 & 46.77 &  57.94 & 86.16 & 96.55 &  91.06 & 90.04 & 63.53 &  74.50 & 86.15 & 96.55 &  91.05 \\
        20\% & 47.86 & 39.53 &  43.30 & 71.43 & 97.83 &  82.57 & 61.66 & 37.77 &  46.84 & 72.41 & 84.00 &  77.78 & 86.29 & 56.45 &  68.25 & 75.44 & 72.88 &  74.14 \\
        30\% & 50.34 & 34.82 &  41.17 & 70.17 & 78.43 &  74.07 & 45.61 & 41.68 &  43.56 & 65.57 & 81.63 &  72.73 & 80.77 & 53.99 &  64.72 & 71.88 & 77.97 &  74.80 \\
        40\% & 45.93 & 24.71 &  32.97 & 64.10 & 60.98 &  62.50 & 40.31 & 20.35 & 27.05 & 57.58 & 69.10 & 62.81 &  72.73 & 53.70 &  61.78 & 68.12 & 79.67 &  73.44 \\
        \bottomrule
    \end{tabularx}
\end{table*}

%% file: Table/clearCompare.tex
\begin{table*}[t]
    \centering
    \small
    \caption{Comparison results with the most related work Clear\cite{clear}. The best results of a certain metric under the same dataset and noise level are highlighted in \textbf{bold}.}
    \resizebox{1\textwidth}{!}{
\begin{tabular}{|l|c|c|c|c|c|c|c|c|c|c|c|c|c|c|c|c|c|c|}
\hline
Type                                                   & \multicolumn{6}{c|}{RE}                                                                             & \multicolumn{6}{c|}{TD}                                                                      & \multicolumn{6}{c|}{IO}                                                                      \\
\hline
\begin{tabular}[c]{@{}l@{}}Noise Level\end{tabular} & \multicolumn{2}{c|}{0\%}            & \multicolumn{2}{c|}{20\%}                            & \multicolumn{2}{c|}{40\%}          & \multicolumn{2}{c|}{0\%}                 & \multicolumn{2}{c|}{20\%}               & \multicolumn{2}{c|}{40\%}          & \multicolumn{2}{c|}{0\%}                   & \multicolumn{2}{c|}{20\%}             & \multicolumn{2}{c|}{40\%}          \\
\hline
Methods                                               & \multicolumn{1}{c|}{Clear} & \multicolumn{1}{c|}{CGBC}         & \multicolumn{1}{c|}{Clear} & CGBC  & \multicolumn{1}{c|}{Clear} & \multicolumn{1}{c|}{CGBC}  & \multicolumn{1}{c|}{Clear} & CGBC  & \multicolumn{1}{c|}{Clear} & \multicolumn{1}{c|}{CGBC}  & \multicolumn{1}{c|}{Clear} & CGBC  & \multicolumn{1}{c|}{Clear} & CGBC  & \multicolumn{1}{c|}{Clear} & \multicolumn{1}{c|}{CGBC}  & \multicolumn{1}{c|}{Clear} & CGBC\\
\hline
P(\%) & \textbf{96.81} & 92.65 & 58.00 & \textbf{71.43} & 50.93 & \textbf{64.10} & 94.30 & \textbf{98.55} & 66.17 & \textbf{72.41} & 50.54 & \textbf{57.58} & 89.81 & \textbf{96.77} & 66.53 & \textbf{75.44} & 52.80 & \textbf{68.12} \\
R(\%) & 96.43 & \textbf{96.92} & 59.18 & \textbf{97.83} & 35.50 & \textbf{60.98} & \textbf{98.41} & 95.77 & 65.19 & \textbf{84.00} & 64.83 & \textbf{69.10} & 91.48 & \textbf{100.00} & 69.30 & 72.88 & \textbf{80.73} & 79.67 \\
F(\%) & \textbf{96.62} & 94.74 & 58.59 & \textbf{82.57} & 41.84 & \textbf{62.50} & 96.31 & \textbf{97.14} & 67.18 & \textbf{77.78} & 56.80 & \textbf{62.81} & 90.64 & \textbf{98.36} & 67.88 & 74.14 & 63.84 & \textbf{73.44} \\
\hline
\end{tabular}
}
\label{tab:clearComparison}
\end{table*}